\documentclass[journal,twoside,web]{ieeecolor}
\usepackage{tmi}
\usepackage{cite}
\usepackage{amsmath,amssymb,amsfonts}
\usepackage{algorithmic}
\usepackage{graphicx}
\usepackage{textcomp}
\usepackage{hyperref}
\usepackage{color}
\usepackage{longtable}
\usepackage{makecell}
\usepackage{multirow}
\usepackage{xcolor}
\usepackage{soul}
\soulregister\cite7
\soulregister\ref7
\soulregister\pageref7

\usepackage{amssymb}
\usepackage{pifont}
\newcolumntype{x}[1]{>{\centering\arraybackslash\hspace{0pt}}p{#1}}

\def\BibTeX{{\rm B\kern-.05em{\sc i\kern-.025em b}\kern-.08em
    T\kern-.1667em\lower.7ex\hbox{E}\kern-.125emX}}
\begin{document}
\title{Deep kernel representations of latent space features for low-dose PET-MR imaging robust to variable dose reduction}
\author{Cameron Dennis Pain*, Yasmeen George, Alex Fornito, Gary F Egan, Zhaolin Chen
\thanks{Manuscript submitted 31 August 2023.
C. D. Pain acknowledges the Australian Government’s support through an Australian Government Research Training Program (RTP) Scholarship.
Z. Chen and G. Egan acknowledge funding support from the Australian Research Council (grants DP210101863 and IM230100002).
A. Fornito was supported by the Sylvia and Charles Viertel Charitable Foundation, the Australian Research Council (ID: FL220100184) and National Health and Medical Research Council (ID: 1197431).
(*corresponding author)}
\thanks{C. D. Pain is with Monash Biomedical Imaging, Monash University, Melbourne, Australia and also with the Department of Electrical and Computer Systems Engineering, Monash University, Melbourne, Australia (e-mail:cameron.pain@monash.edu). }
\thanks{Y. George is with the Department of Data Science and AI, Monash University, Melbourne, Australia}
\thanks{A. Fornito is with the Turner Institute for Brain and Mental Health and School of Psychological Sciences, Monash University, Melbourne, Victoria, Australia and also with Monash Biomedical Imaging, Monash University, Melbourne, Australia}
\thanks{G. F. Egan is with Monash Biomedical Imaging, Monash University, Melbourne, Australia and also with Turner Institute for Brain and Mental Health, Monash University, Melbourne, Victoria, Australia}
\thanks{Z. Chen is with Monash Biomedical Imaging, Monash University, Melbourne, Australia and also with the Department of Data Science and AI, Monash University, Melbourne, Australia}
\thanks{This work has been submitted to the IEEE for possible publication. Copyright may be transferred without notice, after which this version may no longer be accessible.}
} 

\maketitle
\begin{abstract}
Low-dose positron emission tomography (PET) image reconstruction methods have potential to significantly improve PET as an imaging modality. Deep learning provides a promising means of incorporating prior information into the image reconstruction problem to produce quantitatively accurate images from compromised signal. Deep learning-based methods for low-dose PET are generally poorly conditioned and perform unreliably on images with features not present in the training distribution. We present a method which explicitly models deep latent space features using a robust kernel representation, providing robust performance on previously unseen dose reduction factors. Additional constraints on the information content of deep latent features allow for tuning in-distribution accuracy and generalisability. Tests with out-of-distribution dose reduction factors ranging from $\times 10$ to $\times 1000$ and with both paired and unpaired MR, demonstrate significantly improved performance relative to conventional deep-learning methods trained using the same data. Code: \href{https://github.com/cameronPain}{https://github.com/cameronPain}\end{abstract}


\section{Introduction}
\noindent  \IEEEPARstart{P}{ositron} emission tomography (PET) provides a non-invasive means of imaging physiological processes in vivo. Acquiring high quality PET images causes significant radiation exposure to staff and patients in order to generate signal, creating a trade-off between image quality and associated health risks. Image reconstruction is a key component to provide high quality and physically accurate images. low-dose PET image reconstruction is formulated as an ill-posed inverse problem, where poorly sampled data due to reduced radioactivity leads to noisy images with compromised diagnostic quality. A method of compensating for reduced radiation dose is incorporating prior information into the image reconstruction problem to regularise the solution.

\noindent Multi-modality imaging provides additional information inaccessible through a stand-alone modality. While it is standard practice to acquire low-dose CT with PET for physical corrections, PET-MR systems have recently demonstrated the ability to tackle clinical and research problems inaccessible by PET-CT \cite{Chen2018_PETMR_Review}. The excellent soft tissue contrast and high-resolution nature of MR images provides excellent prior information for the PET image processing pipeline with many methods having been developed, including kernel reconstruction techniques \cite{Wang2014_KernelMethod,  8463582, 8118149}  and additive constraints \cite{Mehranian2017_Prior,Raczynski2020,Sudarshan2021}, which involve defining an analytical function relating the two modalities.

\noindent Deep learning has recently demonstrated excellent performance in incorporating population level information into the low-dose PET image processing pipeline \cite{Pain2022_Review} demonstrating impressive performance on dose reduction factors ranging from $\times 2$ \cite{Mehranian2022_LD_multicentre} to beyond $\times 100$\cite{Sudarshan2021}. The data-driven nature of deep learning is well suited for relating two modalities which are unrelated by a physical model, yet share morphological characteristics such as MR, CT and PET. While initial studies provide promising results, a number of pitfalls exist when implementing deep learning-based methods, namely, the generalisability of the performance across varying circumstances experienced in a clinical setting which are not reflected in small cohort datasets. 

\noindent Conventional supervised deep learning approaches are implemented for a dose reduction factor defined by the training data, with clinical workflows subsequently constrained to ensure subsequent data acquisitions are consistently in-distribution, limiting the applicability of the deep-learning method at inference. The flexibility to vary patient dose levels is useful for more radiosensitive cases such as paediatrics and pregnancy. Larger variation in noise levels is observed across patients for low-dose PET owing to the non-linear relationship between signal-to-noise and administered activity for a fixed scan duration \cite{Chang2012_SNR}, making it challenging to ensure acquired data is consistent with training data for very low-levels of administered dose. Current methods designed to handle varying dose levels are either not formulated to leverage high-quality training data, or require exhaustive paired datasets defined across a spectrum of dose reduction factors to reproduce the in-distribution performance of supervised learning methods. Ideally, we aim to implement methods which maintain the in-distribution performance of supervised deep-learning, while providing the flexibility to handle varying dose levels as is offered by self-supervised and conventional regularised image processing methods without burden of additional training data.

\noindent This work presents a novel method for robust deep learning-based processing of low-dose PET images for out-of-distribution dose-reduction factors. Our method provides performance on in-distribution dose levels consistent with conventional supervised learning methods, whilst offering tunablity for improved performance on out-of-distribution dose levels using training data at only a single dose reduction factor. The main novelties and contributions of this work are
\begin{enumerate}
\item Formulating latent-space kernel functions for regularising feature maps to facilitate improved performance across a large spectrum of unseen dose-reduction factors.
\item Applying information constraints on deep latent space features to tune between generalisability across dose reduction factors and in-distribution performance.
\item Applying the proposed method to both paired and unpaired PET-MR data to demonstrate potential for impact beyond situations where both PET and MR are available.
\item Demonstrate performance across $^{18}F$-FDG and $^{18}F$-FDOPA brain images.
\end{enumerate}

\section{Related Works}
\subsection{Deep learning for low-dose PET}

\noindent Deep learning based methods for low-dose to standard-dose PET image synthesis have recently demonstrated promising results. Early works demonstrated the ability to generate high-fidelity images from dose reduction factors beyond $\times 100$ \cite{xu2017,Chen2019_M1,Sudarshan2021} on relatively small cohort studies of healthy subjects. Examples by Xu et al. \cite{xu2017}, and Chen et al.\cite{Chen2019_M1,Chen2021_M1} demonstrated good performance on PET-MR brain data and highlighted the ability of deep learning-based methods for standard-dose image synthesis as well as the utility of coregistered MR.
\noindent Subsequent works implemented generative adversarial networks (GAN) and task-specific losses to synthesise standard-dose images with more realistic textural properties and to preserve pathological features, helping to improve physician graded diagnostic quality. Work by Ouyang et al. \cite{Ouyang2019_M1} incorporated task-specific perceptual loss for preserving pathology associated with neurodegeneration. Similarly, Works by Zhao et al. \cite{Zhao2020_M2} and Zhou et al. \cite{Zhou2020_M2} investigated the cycleGAN implementations for PET brain images and whole body PET brain images respectively for oncological applications. 
\noindent  A key concern of GAN based methods is hallucinations in the synthesised images, causing false positive or false negative cases in clinical situations. Physics guided deep learning implementations aim to provide better data consistency and reduce the required neural network capacity. Physics guided approaches with a learned gradient term demonstrate improved performance over a number of image space implementations with a considerably reduced number of trainable parameters \cite{Gong2019_emnet,Mehranian2019_FBSEM}. Similar work investigated the alternating direction method of multipliers (ADMM) algorithm for integrating deep learning into a constrained optimisation problem \cite{Gong2018_iterative} with improved performance over image space implementations through improved handling of spatially correlated noise.

\subsection{Deep learning for varying dose levels}
\noindent Self-supervised methods have more recently demonstrated promising results in addressing out-of-distribution cases, as they require no standard-dose images in training and often have capacity to fit varying levels of dose on an ad-hoc basis. The deep image prior method \cite{Ulyanov2018_DIP} represents the acquired data with a neural network in a self-supervised manner, where the functional form of the neural network acts as a prior owing to a propensity for fitting image features whilst having a high impedance for noise. Deep image prior methods have been implemented in image space for denoising whole body PET images \cite{Cui2019_DIPImage} and brain images \cite {Sun2021_DIPImage, Hashimoto_2021_4DDIP, Onishi_DIP_Pretrain}. Population-level pre-training was incorporated in work by Cui et al. \cite{Cui2021_population} providing a considerably better initialisation for subject specific fine tuning, allowing for accurate recovery of patient specific uptake with reduced fitting of subject-specific noise. Work by Gong et al. \cite{Gong2018_DIPRecon} formulated the deep image prior for PET image reconstruction using the ADMM algorithm, with subsequent work extending this formulation \cite{Yokota_2019_ICCV, Hashimoto_2022_DIPRecon, Shan2023_PartialDataDIPRecon, Ote2023_DIPlistmode}. Self-supervised methods provide good generalisability on an ad hoc basis, however the performance fails at large dose reduction factors as the functional form of the neural network does not provide a sufficiently strong prior, nor can it leverage information from standard-dose images.

\subsection{Kernel method for low-dose PET}

Conventional regularised image reconstruction approaches, such as MR guided priors \cite{Mehranian2017_Prior,Raczynski2020,Sudarshan2021} and kernel methods \cite{Wang2014_KernelMethod,Wavelet_Kernel} provide robust performance across varying dose levels and explainable results. The regularisation strength of these methods can often be tuned prior to inference using a small number of hyper parameters. The kernel method for PET image reconstruction has been widely explored for regularising PET image reconstruction. Spatio-temporal implementations have been investigated for regularising solutions in the temporal dimension \cite{8463582} and for direct Patlak reconstruction for parametric imaging \cite{8118149}. Graph regularised kernel expectation maximisation was developed to provide tunability between the diagonal and off-diagonal terms to make feasible tuning for lesion recovery \cite{10026307}. Kernel methods have been previously combined with deep learning methods for PET image reconstruction methods. Neural KEM \cite{NeuralKEM} incorporates a kernel matrix for regularisation into the deep image prior method for PET image reconstruction. The deep kernel method \cite{DeepKernelRep} uses a deep learning approach to generating a kernel matrix which can be plugged into a conventional kernel expectation maximisation reconstruction problem. 

\begin{figure*}[t!]
\centering
\includegraphics[width=0.95\textwidth]{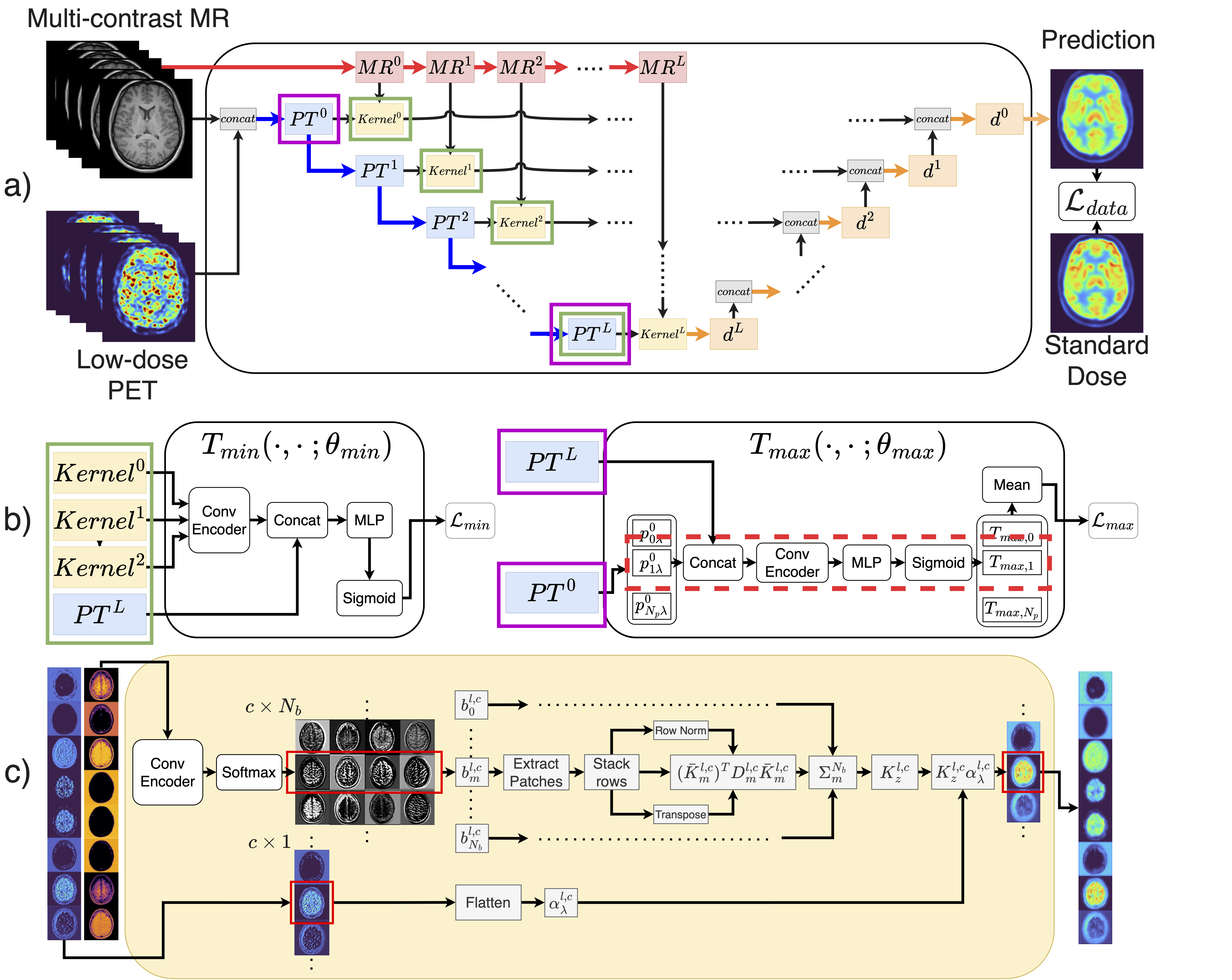}
\caption{ \textbf{a)} The proposed architecture. MR branch features (red) are used to derive latent-space kernel functions. Kernel features (yellow) robust to variations in dose are passed to the decoder. \textbf{b)} Information constraints on latent-space features. Deep PET features (blue) are constrained to minimise mutual information with shallow kernel features. Deep PET representations are constrained to maximise mutual information with patches extracted from shallow PET features. \textbf{c)} Latent space kernel functions are constructed from MR features using a convolutional encoder with a softmax output and constructing symmetric kernel matrices of the form $\mathbf{K}^T\mathbf{D}\mathbf{K}$ where $\mathbf{D}$ is a diagonal matrix.}
\vspace{0.2cm}
\end{figure*}

\section{Notation List}
\begin{itemize}
\item $\boldsymbol{\lambda}$, $\mathbf{z}$, $\mathbf{y}$: PET, MR and sinogram data respectively.
\item $\boldsymbol{v}$: Prior features.
\item $\boldsymbol{\theta}$: Trainable parameters.
\item $\boldsymbol{\alpha}$: Code vector for kernel matrix multiplication.
\item $\boldsymbol{f}_\lambda^l$, $\boldsymbol{f}_z^l$, $\boldsymbol{f}_k^l$: Encoder feature vectors at layer $l$ for PET ($\lambda$), MR($z$) and kernel features ($k$).
\item $\boldsymbol{d}_{z,\lambda}^l$: Decoder feature vectors at decoder layer $l$.
\item $\mathbf{p}_{\lambda}^l$, $p_{i\lambda}^l$, $N_p$: Deep feature vector $\mathbf{f}_{\lambda}^l$ split into patches. The $i^{th}$ patch in $\mathbf{p}_{\lambda}^l$. Total of $N_p$ patches.
\item $f_\lambda^{l,c}$, $f_z^{l,c}$, $f_k^{l,c}$, $d_{z,\lambda}^l$: $c^{th}$ deep feature map contained in deep feature vectors.
\item $\mathcal{F}_{\lambda}^l$, $\mathcal{F}_{z}^l$, $\mathcal{F}_{k}^l$: Distributions from which $\boldsymbol{f}_\lambda^l$, $\boldsymbol{f}_z^l$, $\boldsymbol{f}_k^l$ are sampled.
\item $\Psi(\cdot; \boldsymbol{\theta})$: General mapping from prior features to PET image, parameterised by $\boldsymbol{\theta}$.
\item $k(\cdot,\cdot)$, $\mathbf{K}$: Kernel function, and the kernel matrix defined element-wise by the kernel function.
\item $\Phi(\cdot)$: high-dimensional mapping implicitly defined by kernel function.
\item $N(\cdot; \boldsymbol{\theta})$, $T(\cdot; \boldsymbol{\theta})$: Low-dose to standard-dose neural network and discriminator neural network respectively.
\item $B(\cdot; \boldsymbol{\theta}_{B}^l)$: Deep kernel generating function.
\item $(h^l, w^l, c^l)$, $N_b$, $s$: (width, height, number of feature maps) at layer $l$. Number of softmax features output from $B(\cdot; \boldsymbol{\theta}_{B}^l)$. Stride factor for latent-space kernel matrix.
\item $\psi(\cdot; \boldsymbol{\theta})$, $\phi(\cdot; \boldsymbol{\theta})$: Deep encoder and decoder functions respectively.
\item $\mathcal{N}_{p}(i)$: A square patch of side length $p$ centred at $i$.
\item $\mathcal{L}_{total}, \mathcal{L}_{data}, \mathcal{L}_{I}, \mathcal{L}_{min}, \mathcal{L}_{max}$: Total loss function, L2 norm data-consistency loss, total information constraint, Information minimisation constraint and information maximisation constraint.
\item $\gamma_{max}$, $\gamma_{min}$: hyper-parameters controlling information constraints.
\item $\mathbb{J}_{X,Y}$, $\mathbb{M}_{X}$, $\mathbb{M}_{Y}$: Joint and marginal distributions of random variables $X$ and $Y$.
\item $\mathbb{E}_{\mathbb{P}}[X]$: Expectation value of random variable $X$ distributed according to $\mathbb{P}$ .

\end{itemize}

\section{Background}

\subsection{Parameterising PET images for regularisation}
\noindent Consider a reconstructed PET image $\boldsymbol{\lambda} \in \Lambda$, the corresponding measured signal $\mathbf{y} \in \mathcal{Y}$ and a collection of corresponding prior features $\mathbf{v} \in \mathcal{V}$. An effective way of regularising $\boldsymbol{\lambda}$ when reconstructing sinogram $\mathbf{y}$ is modelling the standard-dose image as a function $\Psi: \mathcal{V} \rightarrow \Lambda$ such that
\begin{equation}
\boldsymbol{\lambda} = \Psi(\mathbf{v}; \boldsymbol{\theta})
\end{equation}
\noindent where $\boldsymbol{\theta}$ is a set of parameters fitting the prior features to the standard-dose image. The function $\Psi$ is chosen with limited ability to model the noise inherent in the low-dose signal and a propensity for modelling standard-dose PET images.\\

\subsection{Kernel functions for modelling PET images}
\noindent Given potentially non-linearly related variables $(x,y) \in \mathcal{X}\times \mathcal{Y}$ kernel methods map input variables to a high dimensional space where similarity across input samples can be measured linearly as a dot product \cite{Hofmann2008_KernelMethodsInMachineLearning}. A function $k:\mathcal{X}\times \mathcal{X} \rightarrow \mathbb{R}$, referred to as a kernel, is defined as 
\begin{equation}
k(x_i, x_j) = \langle \Phi(x_i), \Phi(x_j) \rangle
\end{equation}
\noindent and is used as a similarity measure of inputs $x_i$ and $x_j$, to inform some prediction of corresponding values $y_i$ and $y_j$, where $\Phi$ is the mapping to higher dimensions.

Defining the kernel function explicitly using a symmetric function avoids defining $\Phi$ and performing calculations in high-dimensional space. The choice of kernel function implicitly defines the high-dimensional mapping function. Table 1 lists commonly used kernel functions.

\begin{table}[t!]
\centering
\caption{\footnotesize{Commonly defined kernel functions. In this table, $\sigma$ and $n$ represent free variables.}}
\begin{tabular}{m{2.6cm} >{\centering\arraybackslash}m{5.cm}}
\hline
name &  function\\
\hline
Linear & $k(x_i, x_j) = x_i^T x_j$   \\
Gaussian Radial Basis Function & $k(x_i, x_j) = exp(-\tfrac{1}{2\sigma} ||x_i - x_j||^2)$\\
Polynomial & $k(x_i, x_j) = (x_i^Tx_j + \sigma)^n$ \\
\hline
\end{tabular}
\vspace{-0.4cm}
\end{table}

\noindent Given the task posed by equation (1), the kernel method is a natural choice for modelling low-dose PET images. As formulated by Wang and Qi \cite{Wang2014_KernelMethod}, equation (1) can be reformulated with a kernel function such that

\begin{equation}
\lambda_i =  \sum_{j} \alpha_j \Phi^T(v_j)\Phi(v_i)
\end{equation}

\noindent which reformulates equation (1) into
\begin{equation}
\boldsymbol{\lambda} = \mathbf{K}\boldsymbol{\alpha}
\end{equation}
\noindent where the kernel matrix $\mathbf{K}$ is derived from an explicitly defined function of $\mathbf{v}$, where prior features such as coregistered MR images or a temporally summed PET image are often used, and coefficient vector $\boldsymbol{\alpha}$ is iteratively updated to fit the measured signal. \\

\subsection{Deep learning for modelling PET images}
\noindent Deep learning methods for low-dose PET use a neural network $N:\mathcal{V} \rightarrow \Lambda$ to model the function $\Psi$ in equation (1) such that
\begin{equation}
\boldsymbol{\lambda} = N(\mathbf{v}; \boldsymbol{\theta})
\end{equation}
\noindent where the prior features $\mathbf{v}$ are often MR in combination with low-dose reconstructed PET or PET sensor space data, and the parameters $\boldsymbol{\theta}$ are optimised through training. Modelling standard-dose PET images as in equation (5) can facilitate post-processing methods formulated as
\begin{align}
\boldsymbol{\lambda}_{ld} &= \underset{\boldsymbol{\lambda}}{\text{argmin}} -log P(\mathbf{y}|\boldsymbol{\lambda}) \\
\boldsymbol{\lambda} &= N(\mathbf{v}; \boldsymbol{\theta}),\  \lambda_{ld} \in \mathbf{v} 
\end{align}
where $P(\mathbf{y}|\boldsymbol{\lambda})$ is the conditional probability distribution of $\mathbf{y}$ given $\boldsymbol{\lambda}$ and $\boldsymbol{\lambda}_{ld}$ is the reconstructed low-dose PET image which is included as an element of the prior vector $\mathbf{v}$. Alternatively, it may be incorporated into an expectation maximisation reconstruction where the PET image is explicitly modelled with a neural network as
\begin{equation}
\hat{ \boldsymbol{\lambda} }= \underset{\boldsymbol{\lambda}}{\text{argmin}} -log P(\mathbf{y}|\boldsymbol{\lambda}),\  \boldsymbol{\lambda} = N(\mathbf{v}; \boldsymbol{\theta})
\end{equation}

\noindent Approaches with pre-trained and fixed neural network parameters are used for the approach outlined in equation(8) where the network input $\mathbf{v}$ is iteratively updated to solve equation (8) \cite{Gong2018_iterative}. Similarly, supervised learning for post-processing methods as outlined in equation (7) have been widely implemented \cite{Chen2019_M1,Chen2021_M1,Ouyang2019_M1, Zhou2020_M2, Zhao2020_M2}. Unsupervised approaches based on the deep image prior method \cite{Ulyanov2018_DIP} have been developed for solving equations (7) \cite{Hashimoto2021_DIP} and equation (8) \cite{Gong2018_DIP, Ote2023_DIPlistmode}.\\

\subsection{Kernel methods vs deep learning}
\noindent Kernel methods are incorporated into conventional model-based EM reconstruction algorithms in a straight-forward manner, providing convergence and robust generalisability. However, these methods can lack capacity to model features such as PET specific lesions which may not be present in MR contrast used to define the kernel function. Deep learning methods provide exceptional modelling capacity, producing high-fidelity PET images beyond that achievable with kernel methods, however consistent performance on out-of-distribution data is not guaranteed.

\section{Method}

\noindent This work presents a latent-space kernel method which is tunable at training to determine noise control and modelling capacity, and is robust across out-of-distribution dose reduction factors at inference. Figure 1 depicts the layout of the proposed network. Low-dimensional MR derived kernel functions are embedded into the latent space of a neural network to regularise feature maps. The proposed implementation relies on reliably in-distribution MR images at inference for effective regularisation in the presence of previously unseen noise levels in the PET. We additionally add information constraints to the deep latent space features to control what features are encoded into deep layers of the network. By constraining deep latent features to encode global level information and preference shallow kernel representations over deep ones, we steer the network towards learning latent representations more robust to previously unseen reduced dose levels.

\subsection{Deep kernel network}
\noindent We term the neural network with kernel functions embedded into the latent space the deep kernel network. For a low-dose PET-MR pair $(\boldsymbol{\lambda}_{ld}, \textbf{z}) \in \Lambda \times \mathcal{Z}$, we model equation (1) with a deep kernel network $ N :  \Lambda \times \mathcal{Z} \rightarrow \Lambda$ where
\begin{equation}
\boldsymbol{\lambda} = N(\boldsymbol{\lambda}_{ld}, \textbf{z}; \boldsymbol{\theta})
\end{equation}
\noindent The deep kernel network consists of two distinct encoding branches: 1) the MR encoding branch (denoted subscript $z$) and 2) the PET encoding branch (denoted subscript $\lambda$). Each encoding branch is comprised of a series of $L$ encoders $\psi^{l}_{\lambda}(\cdot; \boldsymbol{\theta}_{\lambda}^l)$ and $\psi^{l}_{z}(\cdot; \boldsymbol{\theta}_{z}^l)$ for $l \in [L]$ where we use the notation $[L] = \{0, .., L\}$. PET and MR encoders learn latent representations $\{\mathbf{f}_\lambda^l \in \mathcal{F}_\lambda^l \subset \mathbb{R}^{h_l \times w_l \times c_l}: l \in [L]\}$ and $\{\mathbf{f}_z^l \in \mathcal{F}_z^l \subset \mathbb{R}^{h_l \times w_l \times c_l}: l \in [L]\}$ respectively, where $h_l, w_l$ and $c_l$ are the height, width and number of feature maps at layer $l$. Feature vector $\mathbf{f}_z^l$ from the MR branch is used to define a collection of $c^l$ kernel functions $\mathbf{K}^{l,c}_z$ for $c \in [c^l]$ and feature vector $\mathbf{f}_\lambda^l$ from the PET branch is used to define code vectors $\alpha_\lambda^{l,c}$ for $c \in [c^l]$, from which a collection of kernel features $f_{k}^{l,c} \in \mathbf{f}_k^{l}$ are defined as
\begin{equation}
f_{k}^{l,c} = \mathbf{K}^{l,c}_z \alpha^{l,c}_{\lambda}
\end{equation}
\noindent  where $\mathbf{K}^{l,c}_z$ is a function of MR and code vector $\alpha^{l,c}_{\lambda}$ is a function of PET. A single Unet-style decoder branch with serial layers $\mathbf{d}_{z,\lambda}^l = \phi_{z,\lambda}^{l}(\mathbf{f}_{k}^{L-l}, \mathbf{f}_z^{L-l}, \mathbf{d}_{z,\lambda}^{l-1}; \mathbf{\theta}_d^l)$ for $l \in [L]$ is used to map encoded features to the target image domain.

The structure of $\mathbf{K}_z^{l,c}$ ultimately defines the level of regularisation we introduce. If we consider $\mathbf{K}_{z}^{l,c}$ as diagonal matrices, then we effectively introduce no explicit MR guidance into the kernel features $\mathbf{f}_k^l$. Similarly, if $\mathbf{K}_z^{l,c}$ are dense matrices (mostly non-zero elements), then we enforce $\mathbf{f}_{k}^{l}$ to have a strong dependence on MR. We utilise this formulation to tune $\mathbf{K}_z^{l,c}$ in order to provide a trade-off between performance across widely varying dose levels and peak performance in-distribution.

\subsection{Kernel method for modelling deep latent features}
\noindent Kernel functions as shown in equation (10) are defined for each layer $l$ using a learned encoder $B^l(\cdot ; \boldsymbol{\theta}_{B}^l) : \mathcal{F}_z^l \rightarrow (0,1)^{h_l \times w_l \times c_l \times N_b}$ as
\begin{equation}
\mathbf{b}^l = B^l(\mathbf{f^l_z}; \boldsymbol{\theta}_B^l)
\end{equation}
\noindent where $b^{l,c} \in \boldsymbol{b}^{l}$ is a collection of $N_b$ feature maps used to construct kernel function $\boldsymbol{K}_z^{l,c}$. We use a softmax output such that $\sum_{m}^{N_b} b^{l,c}_{m} = 1$, where $N_b$ is the number of the prior features used for calculating a latent-space kernel matrix. The encoder $B^l(\cdot; \boldsymbol{\theta}_B^l)$ acts to transform latent features $\mathbf{f}_z^l$ into features $\mathbf{b}^l$ optimised for generating kernel functions. This allows $\mathbf{f_z^l}$ to be simultaneously optimised for deeper network layers and kernel functions, at the cost of added computational burden. The softmax function acts to normalise the elements of the resultant kernel function, improving convergence during training. For each $m \in [N_b]$, a matrix $\mathbf{\bar{K}}_m^{l,c}$ is then defined as
\begin{equation}
k^{l,c}_{i,j,m} = \begin{cases} 
      b^{l,c}_{j,m} &  j \in \mathcal{N}_{p}(i)\\
      0 & \text{elsewhere}
      \end{cases}
\end{equation}
\noindent where $\mathcal{N}_{p}(i) \in \mathbb{R}^{p \times p}$ defines a square patch in $\mathbf{b}^{l,c}$ with side length $p$ centred at $i$. We define latent-space kernel matrices from equation (10) as
\begin{equation}
\mathbf{K}^{l,c}_z = \sum_m^{N_b}(\boldsymbol{\bar{K}}^{l,c}_{m})^T \boldsymbol{D}^{l,c}_m \boldsymbol{\bar{K}}^{l,c}_{m}
\end{equation}
Where $\boldsymbol{D}^l_m$ is a diagonal matrix used for normalisation with elements defined as
\begin{equation}
d_{i,i,m}^{l,c} =\dfrac{1}{\sum_j k^{l,c}_{i,j,m}}
\end{equation}
\noindent Code vectors $\alpha_{\lambda}^{l,c}$ from equation (10) are defined as flattened PET branch features $\alpha_{\lambda}^{l,c} = f_{\lambda}^{l,c}$. \noindent There are a number of key differences between image space and latent-space kernel functions which must be considered. Conventional methods perform a single calculation of the image-space kernel function and save a sparse matrix using a $k$ nearest neighbours algorithm for subsequent image reconstruction. Our method requires calculation of kernel functions for each forward pass during training and the storage of gradients for updating parameters. While a $k$ nearest neighbours can facilitate non-local connections between kernel elements, a patch-based approach allows us to define a sparse matrix without explicitly calculating each kernel element, saving computational power and memory requirements during training. Additionally, because our method derives kernel functions from learned feature space representations and rather than fixed image space prior features, and we utilise a collection of kernel functions rather than a single kernel function, it is not clear what effect the choice of kernel function has on performance. Equation (13) is equivalent to a linear kernel as listed in table 1, however other symmetric functions may be considered.

\subsection{Stride latent-space kernel function}
\noindent To reduce the dimensionality of $\mathbf{K}^{l,c}_z$ for more time and memory efficient training, a matrix $\mathbf{\bar{K}}^{l,c,s}_z$ with stride factor $s$ is derived from 
\begin{equation}
k^{l,c,s}_{i',j,m} = \begin{cases} 
      b^{l,c}_{j,m} &  j \in \mathcal{N}_{p}(i'), \text{where}\ i' = si\\
      0 & \text{elsewhere}
      \end{cases}
\end{equation}
and is used to generate a more computationally efficient stride latent-space kernel matrix
\begin{equation}
\mathbf{K}^{l,c,s}_z = \sum_m^{N_b}(\boldsymbol{\bar{K}}^{l,c,s}_{m})^T \boldsymbol{D}^{l,c,s}_m \boldsymbol{\bar{K}}^{l,c,s}_{m}
\end{equation}
with robust PET features calculated in the same way as equation (10). A stride size $s$ reduces the number of elements calculated over by a factor of $1/s^2$. The matrix $\mathbf{\bar{K}}^{l,c,s}_{m}$ become rectangular in this instance, mapping features to an intermediate lower dimensional representation providing reduced computational requirements, impedance to noise, yet reduced modelling capacity.

\subsection{Information constraints on deep PET features}
\noindent Information constraints on deep layers are imposed to control the nature of the information encoded in deep layers of the network. We hypothesise that shallow kernel features provide representations more robust to previously unseen variations in dose reduction, therefore we aim to penalise the network for encoding information deep into the network which may be adequately represented in shallow layers.
Deep PET branch features are left to encode that information which is unable to be modelled by shallow kernel representations. We achieve this by applying two constraints to deep PET features.\\

\subsubsection{Information minimisation constraint}
\noindent For latent-space kernel features $\{ \mathbf{f}_{k}^l : l \in [L]  \}$ and PET features $\{ \mathbf{f}_{\lambda}^l : l \in [L] \}$ a discriminator network $T_{min}(\cdot, \cdot; \boldsymbol{\theta}_{min}): \mathcal{F}_{k}^l \times \mathcal{F}_\lambda^L \rightarrow (0,1)$ predicts whether $(\mathbf{f}_k^l, \mathbf{f}_\lambda^L)$ is sampled from the joint distribution $\mathbb{J}_{\mathcal{F}_k^l, \mathcal{F}_\lambda^L}$ or independently sampled from the marginals $\mathbb{M}_{\mathcal{F}_{k}^l}$ and $\mathbb{M}_{\mathcal{F}_\lambda^L}$. The mutual information minimisation constraint 
\begin{equation}
\mathcal{L}_{min} = -\mathbb{E}_{\mathbb{J}_{\mathcal{F}_{k}^l \times \mathcal{F}_\lambda^L}} [\log(1 - T_{min}(\mathbf{f}_k^l, \mathbf{f}_\lambda^L; \boldsymbol{\theta}_{min}))]
\end{equation}
\noindent is added to the training loss function of the PET encoding branch only. The term $\mathcal{L}_{min}$ is minimised when the mutual information between shallow kernel features $\mathbf{f}_k^l$ and deep PET features $\mathbf{f}_\lambda^L$ is minimised. As the MR encoder is not updated with respect to this loss, deep PET features $\mathbf{f}_\lambda ^L \in \mathcal{F}_\lambda^L$ are penalised for encoding information available to shallow kernel features, effectively forcing the network to preference shallow representations when possible. Shallow kernel features are likely less prone to corruption by reduced dose reduction factors owing to the averaging effects of the shallow kernel functions, providing better generalisability at the expense of modelling capacity.

\begin{figure*}[h!]
\centering
\includegraphics[width=0.95\textwidth]{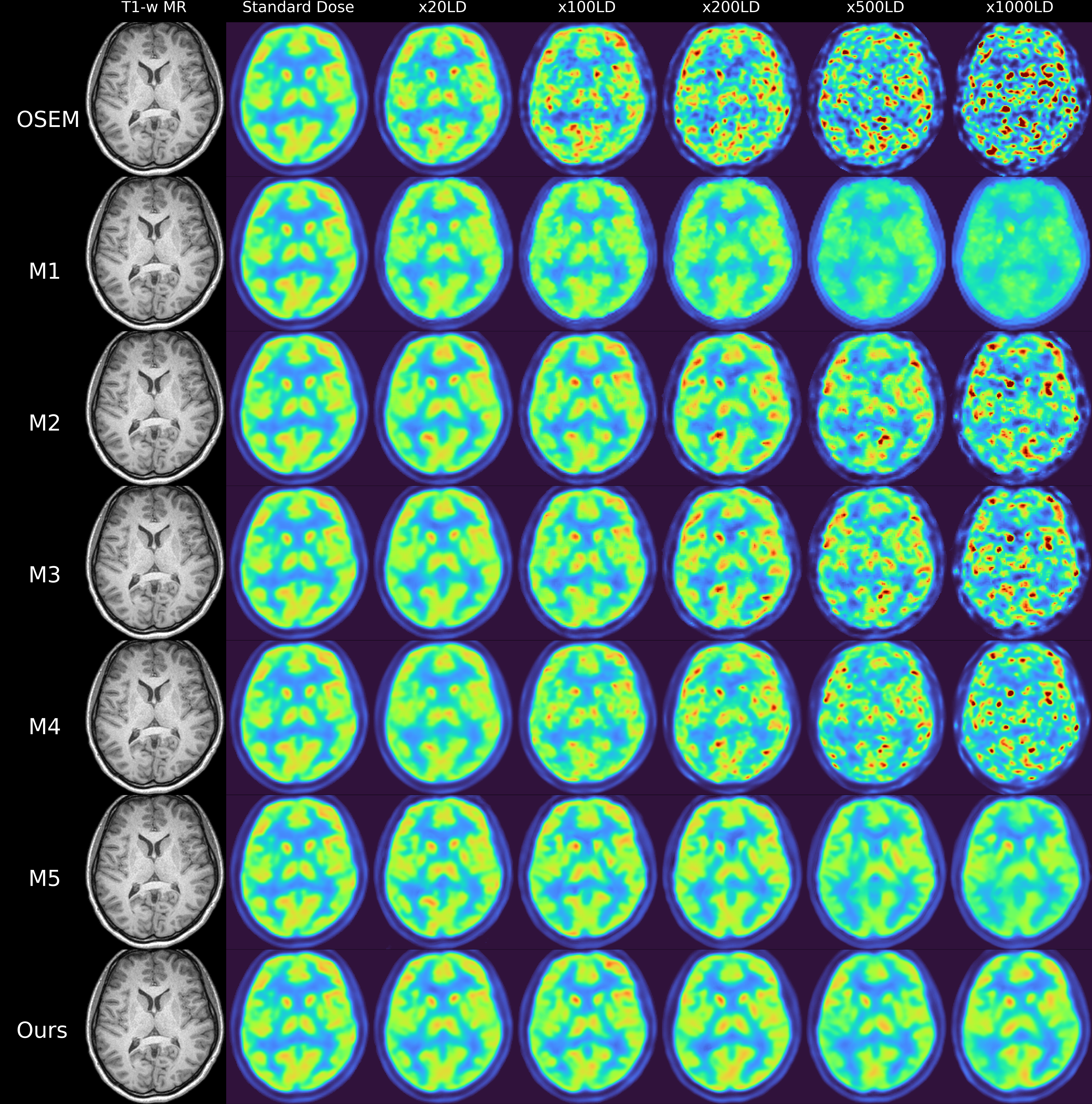}
\caption{Comparison of our method with benchmarks from the literature trained on $\times 20$ low-dose PET and paired MR images. The proposed method uses kernel patch sizes of $1,8,32,32,32$ for dose reduction factors $\times 20$,$\times 100$,$\times 200$,$\times 500$,$\times 1000$ respectively. We apply information constraints $\gamma_{max} = 0.01$ and $\gamma_{min} = 0.001$ for dose reduction factors of $\times 500$ and $\times 1000$.}
\end{figure*}


\begin{figure*}[h!]
\centering
\includegraphics[width=0.95\textwidth]{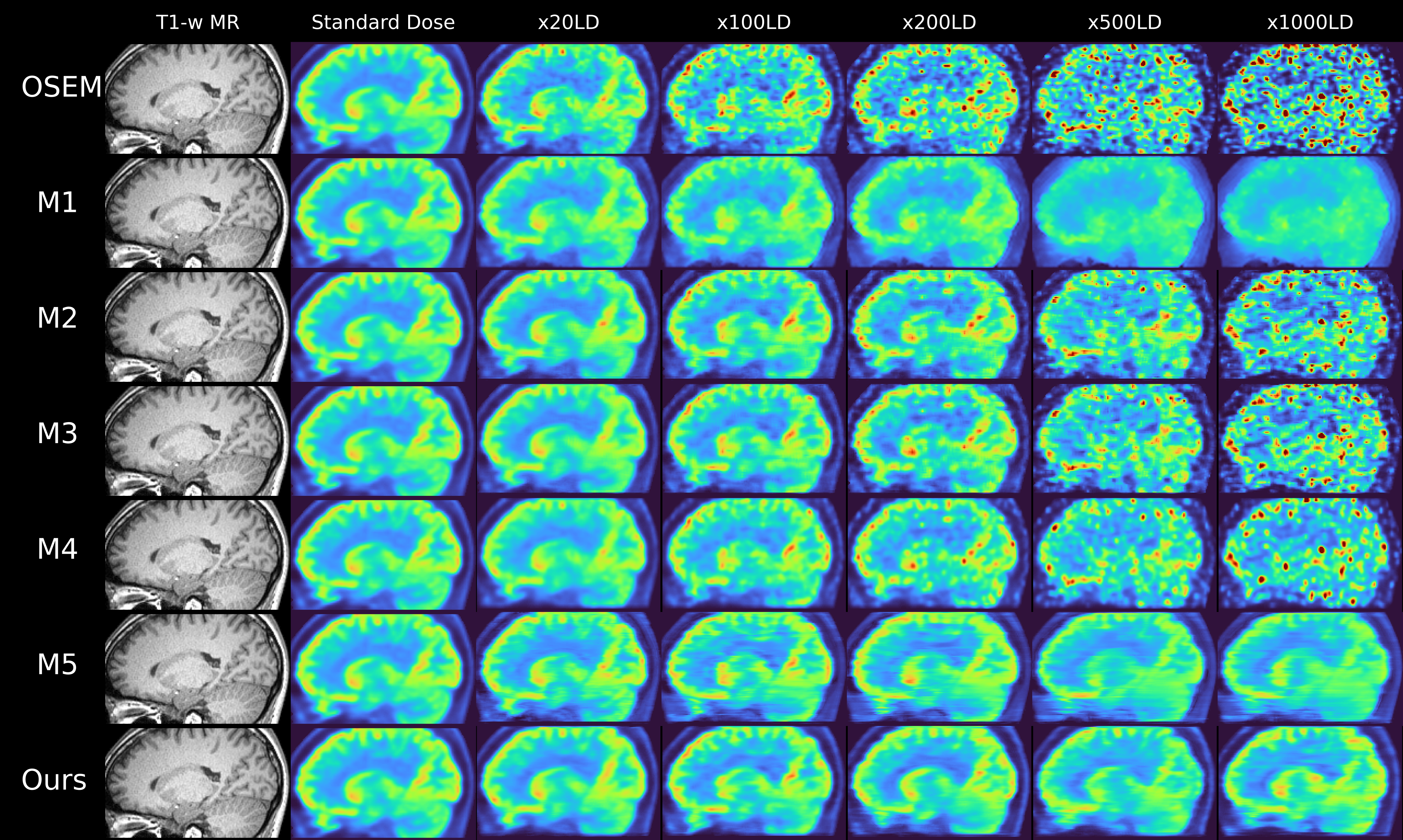}
\caption{Comparison of our method with benchmarks from the literature trained on $\times 20$ low-dose PET and paired MR images. The proposed method uses kernel patch sizes of $1,8,32,32,32$ for dose reduction factors $\times 20$,$\times 100$,$\times 200$,$\times 500$,$\times 1000$ respectively. We apply information constraints $\gamma_{max} = 0.01$ and $\gamma_{min} = 0.001$ for dose reduction factors of $\times 500$ and $\times 1000$.}
\end{figure*}

\subsubsection{Information maximisation constraint}
In a manner similar to that of Hjelm et al. \cite{hjelm2019_DeepInfoMax}, we constrain deep PET features to encode global information. Shallow PET features $\mathbf{f}_\lambda^l$ are split into a collection of $N_p$ patches $\mathbf{p}_{\lambda}^l \in \mathcal{F}_{p,\lambda}^l$. Each patch $p_{\lambda}^l \in \mathbf{p}_{\lambda}^l$ is concatenated channel-wise with deep PET features $\mathbf{f}_\lambda ^L$ to form pairs $(p_{\lambda}^l, \mathbf{f}_{\lambda}^L)$. Discriminator network $T_{max}(\cdot, \cdot; \boldsymbol{\theta}_{max}) :  \mathcal{F}_{p,\lambda}^l \times \mathcal{F}_{\lambda}^L \rightarrow (0,1)$ predicts whether $(p_\lambda^l, \mathbf{f}_\lambda^L)$ is sampled from the joint distribution $\mathbb{J}_{\mathcal{F}_{p,\lambda}^l, \mathcal{F}_\lambda^L}$ or independently sampled from the marginals $\mathbb{M}_{ \mathcal{F}_{p,\lambda}^l}$ and $\mathbb{M}_{\mathcal{F}_\lambda^L}$. The mutual information maximisation constraint 

\begin{equation}
\mathcal{L}_{max}(\mathbf{p}_{\lambda}^l),\mathbf{f}_\lambda^L)  = \frac{-1}{N_p} \sum_{p_{\lambda}^l \in \mathbf{p}_{\lambda}^l}  \mathbb{E}_{\mathbb{J}_{ \mathcal{F}_{p,\lambda}^l \times \mathcal{F}_\lambda^L}}[\log (T_{max}(p_{\lambda}^l, \mathbf{f}_\lambda^L; \boldsymbol{\theta}_{max})]
\end{equation}

\noindent is added to the training loss function to require global PET features are propagated to deep layers of the network penalising the network for learning representations of small anatomical locations easily corrupted by reduced dose.

\subsubsection{Total information constraint} The information content of the deep features is constrained using the loss term \begin{equation}
\mathcal{L}_{I} = \gamma_{max} \mathcal{L}_{max} + \gamma_{min} \mathcal{L}_{min}
\end{equation}
\noindent when training the PET encoders $\{ \psi_\lambda^i(\cdot; \boldsymbol{\theta}_{\lambda}^l):i\in [L]\}$ where $\gamma_{max}$ and $\gamma_{min}$ are hyper-parameters which control the strength of the information constraints.

\subsection{Training strategy}
\noindent The deep kernel network is trained to minimise the loss
\begin{equation}
\mathcal{L}_{total} = \mathcal{L}_{data} + \mathcal{L}_I
\end{equation}

\noindent where $\mathcal{L}_{data}$ is the L2 norm between the predicted output and the standard-dose PET reference and $\mathcal{L}_I$ is the total information constraint as in equation (19). During backpropagation, gradients from the information constraint loss term are only propagated through the PET branch.
 The discriminator $T_{max}(\cdot, \cdot;  \boldsymbol{\theta}_{max})$  used in the information maximisation constraint has a cooperative objective function with the deep kernel network and is therefore trained simultaneously with the deep kernel network to minimise the loss function in equation (20). 
The discriminator $T_{min}(\cdot, \cdot;  \boldsymbol{\theta}_{min})$ used in the information minimisation constraint has an adversarial objective function with the deep kernel network and is trained in an alternating manner to maximise equation (17). For training both  $T_{min}(\cdot, \cdot;  \boldsymbol{\theta}_{min})$ and $T_{max}(\cdot, \cdot;  \boldsymbol{\theta}_{max})$, samples from the marginal distributions are taken by rolling inputs along the batch axis and maximising equations (17) and (18) respectively.

\section{Experiments}
\noindent The performance of the proposed method is evaluated in two contexts: 1) Simultaneously acquired PET-MR, and 2) unpaired PET and MR images. These contexts serve to evaluate the proposed method in an ideal situation of simultaneously acquired PET-MR, and the PET only case where we use a template MR image common across all subjects. We present an ablation study on the latent-space kernel functions, comparing the performance of linear and gaussian kernel functions, and the effect of varying kernel function patch-size for varying levels of regularisation. Finally, we investigate the effect of information constraints applied to varying layers of the network and for varying hyper-parameter strengths. 

\subsection{Data}
\noindent Data acquisition was performed after receiving ethics approval by the institution. Simultaneously acquired MR images included ultra-short echo time, T1 MPRAGE and T2-SPC. PET images were reconstructed into $2.09 \times 2.09 \times 2.03\ mm^3$ voxels with Siemens e7 tools software using an ordered subsets expectation maximisation algorithm with 21 iterations, 3 subsets and point spread function modelling. Attenuation correction was applied using $\mu$-maps derived from UTE MR images. Low-dose PET data was simulated by randomly sampling acquired list-mode data. PET images were normalised to activity concentration units of $kBq/ml$ and scaled by the dose reduction factor.

\subsubsection{$^{18}F$-FDG PET-MR Brain Images}
\noindent A cohort of 28 healthy individuals (mean age 19.6 years, range 18-22, 21 females) were scanned on a 3T Siemens Biograph mMR system. Subjects were administered approximately 250 MBq of $^{18}$F-FDG. Standard dose PET images were reconstructed from 30 minutes of data acquired 60 minutes post-administration. Low-dose PET data was derived at dose reduction factors of $\times 20$, $\times 100$, $\times 200$, $\times 500$ and $\times 1000$. 

\subsubsection{$^{18}F$-FDOPA PET-MR Brain Images}
\noindent A cohort of 25 individuals (mean age 21.6 years, range 18-25, 8 females) were scanned on a 3T Siemens Biograph mMR system. Subjects were administered approximately 150 MBq of $^{18}$F-FDOPA. Standard dose PET images were reconstructed from 46 minutes of data acquired 30 minutes post-administration. Low-dose PET data was derived at dose reduction factors of $\times 10$, $\times 20$, $\times 50$, $\times 100$ and $\times 200$ producing a peak signal to noise ratio (PSNR) equivalent to the low-dose $^{18}F$-FDG datasets due to a relatively lower tracer uptake.

\begin{figure*}[h!]
\centering
\includegraphics[width=0.95\textwidth]{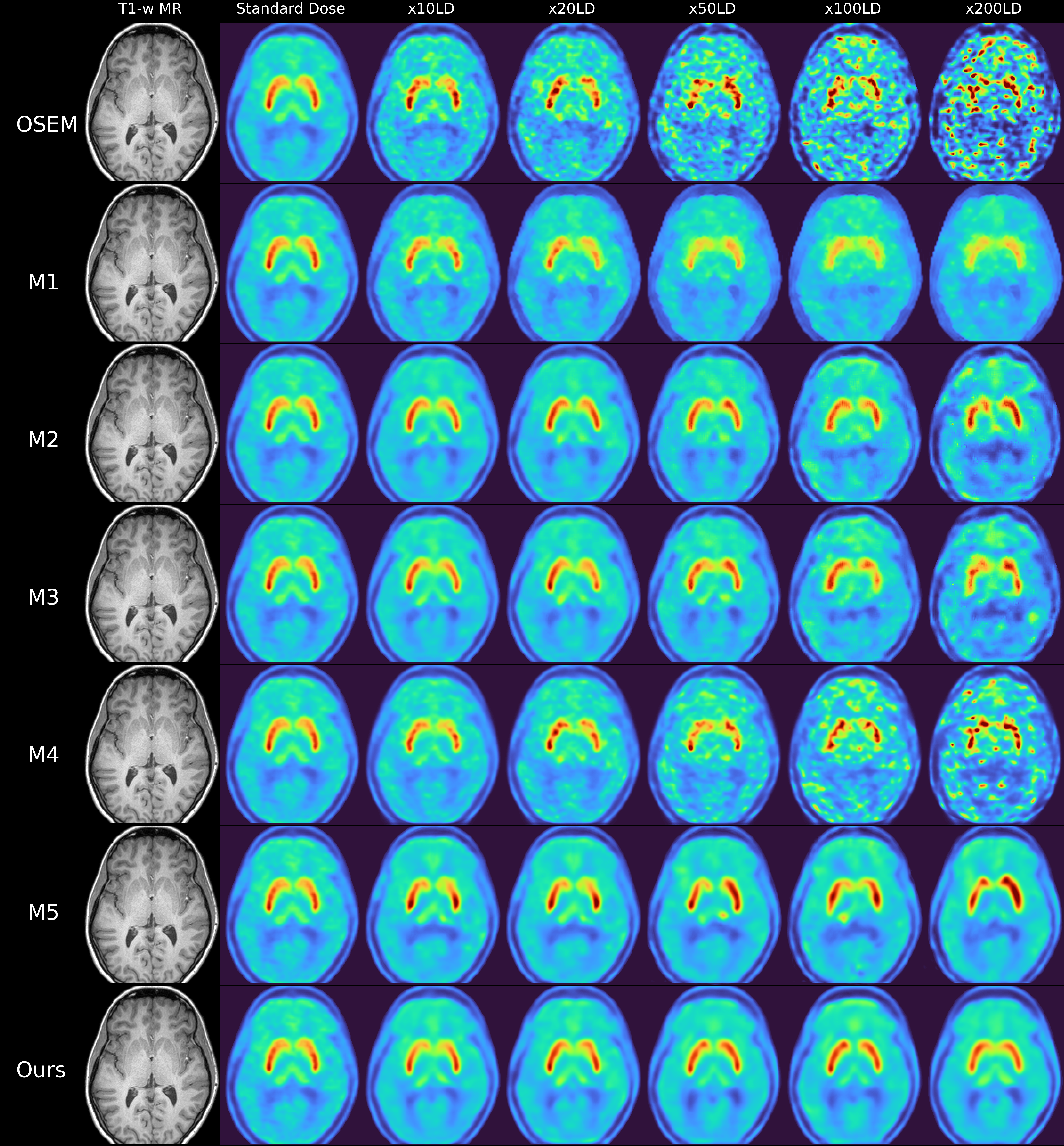}
\caption{Comparison of our method with benchmarks from the literature trained on $\times 10$ low-dose [F18]-FDOPA and paired MR images. The proposed method uses kernel patch sizes of $1,8,32,96,96$ for dose reduction factors $\times 10$,$\times 20$,$\times 50$,$\times 100$,$\times 200$ respectively. We apply information constraints $\gamma_{max} = 0.01$ and $\gamma_{min} = 0.001$ for dose reduction factors of $\times 200$ and $\gamma_{max} = 0.1$ and $\gamma_{min} = 0.001$ for $\times 200$.}
\end{figure*}


\begin{figure*}[h!]
\centering
\includegraphics[width=0.95\textwidth]{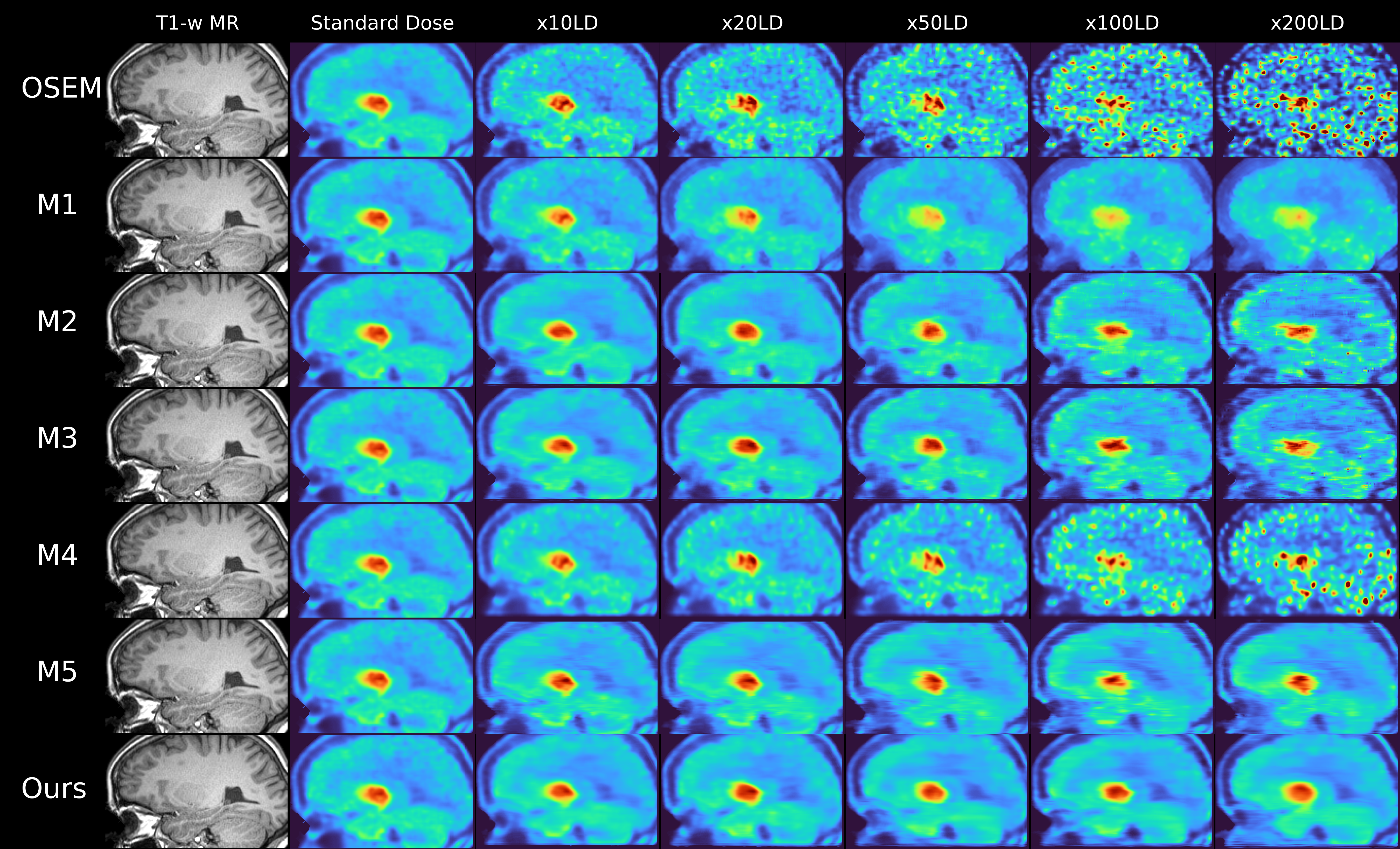}
\caption{Comparison of our method with benchmarks from the literature trained on $\times 10$ low-dose [F18]-FDOPA and paired MR images. The proposed method uses kernel patch sizes of $1,8,32,96,96$ for dose reduction factors $\times 10$,$\times 20$,$\times 50$,$\times 100$,$\times 200$ respectively. We apply information constraints $\gamma_{max} = 0.01$ and $\gamma_{min} = 0.001$ for dose reduction factors of $\times 200$ and $\gamma_{max} = 0.1$ and $\gamma_{min} = 0.001$ for $\times 200$.}
\end{figure*}

\subsubsection{Simultaneously acquired PET-MR}
We used ANTS coregistration software \cite{ANTS2009} to perform image coregistration tasks. A rigid registration was used to coregister standard-dose PET and T2 MR to the T1 MR image with $1 \ mm^3$ isotropic voxels. The transformation calculated for the standard-dose PET was applied to low-dose the PET image for optimal coregistration.

\subsubsection{Unpaired PET-MR}
A coregistered T1 and T2 weighted MR pair from a nominal subject was used as an unpaired coregistration template for PET images from separate subjects. A series of rigid, affine and deformable registrations were used to coregister PET images to the selected template, with the transformation calculated for standard-dose PET used for coregistering low-dose PET. 

\subsection{Evaluation Metrics}
\noindent Results were quantified according to the peak signal to noise ratio (PSNR) and structural similarity index measure (SSIM) defined as
\begin{align}
MSE(x,y) &= \dfrac{1}{N} \sum_{i = 0}^N (x_i - y_i)^2\\
PSNR(x,y) &= 20 \log_{10}\Big(\dfrac{\max(y)}{MSE(x,y)}\Big)\\
SSIM(x,y) &= \dfrac{(2\mu_x \mu_y + c_1)(2\sigma_{xy} + c_2)}{(\mu_x^2 + \mu_y^2 + c_1)(\sigma_x^2 + \sigma_y^2 + c_2)}
\end{align}
where $x$ and $y$ are predicted and ground truth images respectively. For the SSIM calculation, $\mu_x$ and $\mu_y$ are the mean values in the SSIM window, $\sigma_x$, $\sigma_y$ and $\sigma_{xy}$ are the variances and covariance in the SSIM windows, $c_1 = 0.01R$ and $c_2=0.03R$ where $R = \max(y)$. A window size of 7 was used for all SSIM calculations with the reported values being the mean SSIM value for an image.

Regional uptake was quantitatively analysed by co-registering output images to the Harvard-Oxford subcortical brain atlas sourced from FSL\cite{Jenkinson2012_FSL}, using A series of rigid, affine and deformable registrations. The mean normalised bias and variance in uptake relative to standard-dose PET was calculated for the gray matter, white matter, thalamus, putamen and caudate. 


\subsection{Neural Network Implementation}
\noindent The proposed method is implemented as a $2.5D$ approach, with approximately 150 axial slices taken from each subject. Inputs consist of 5 contiguous axial slices from each of \color{black} the three input modalities, concatenated along the feature axis for input dimensions of $192\times192\times15$, with each dimension representing anterior-posterior, left to right, and the 5 contiguous slices from low-dose PET, T1 weighted MR, and T2 weighted MR. The output from the neural network is $192\times192\times1$ in shape, which is trained in a supervised manner with a standard-dose PET slice target, corresponding to the centre of the 5 contiguous input slices. 

\noindent The network shape used in this work is consistent with a Unet. Encoding layers for each branch  are a series of two convolutional blocks each consisting of $3\times3$ convolutional filters, batch normalisation and ReLU activation in series, followed by a $2 \times 2$ max pooling. Decoding layers consist of a bilinear upsampling and concatenation with shallow features in series with two convolutional blocks. The encoder and decoders both consist of 6 layers. Functions $B^l(\mathbf{\cdot}; \boldsymbol{\theta}_B)$ are two convolutional blocks in series with a softmax along the feature axis and $N_b = 4$ softmax features.

\noindent The information maximisation constraint discriminator $T_{max}(\cdot, \cdot; \boldsymbol{\theta}_{max})$ takes $24\times 24$ patches extracted from features $\boldsymbol{f}_{\lambda}^0$  concatenated with features $\boldsymbol{f}_{\lambda}^3$ as input. The network architecture is a series of three convolutional blocks consisting of $3\times3$ convolutional filters, batch normalisation, ReLU activation and 2D maximum pooling followed by a multi layer perceptron using three layers with 16, 8 and 1 neurons and a sigmoid activation function at the output.
\noindent The information minimisation constraint discriminator $T_{min}(\cdot, \cdot; \boldsymbol{\theta}_{min})$ takes kernel features $\boldsymbol{f}_k ^ l$ for $l=0,1,2$ and  $\boldsymbol{f}_{\lambda}^5$ as input. The network archtitecture is a convolutional encoder 
\color{black} with the same architecture as the encoder used for the PET branch of the deep kernel network. Kernel features $\boldsymbol{f}_k ^ 0$ are 
\color{black}used as inputs, with $\boldsymbol{f}_k ^ l$ for $l=1,2$ concatenated with the inputs to subsequent layers and $\boldsymbol{f}_{\lambda}^5$ concatenated at the final convolutional layer. A multi layer perceptron with the same architecture as for the $T_{max}(\cdot, \cdot; \boldsymbol{\theta}_{max})$ discriminator is used to map convolutional features to the output.

\color{black}
\noindent Free parameters were updated using the Adam optimisation algorithm with initial learning rate $l = 0.0001$, exponential decay factor of $0.99^{N/100}$ where $N$ is the current number of batches trained over, and a batch size of 20. Training was performed using two Nvidia A40 with a total training time of approximately $5$ hours per neural network.

\begin{figure*}[h!]
\centering
\includegraphics[width=0.95\textwidth]{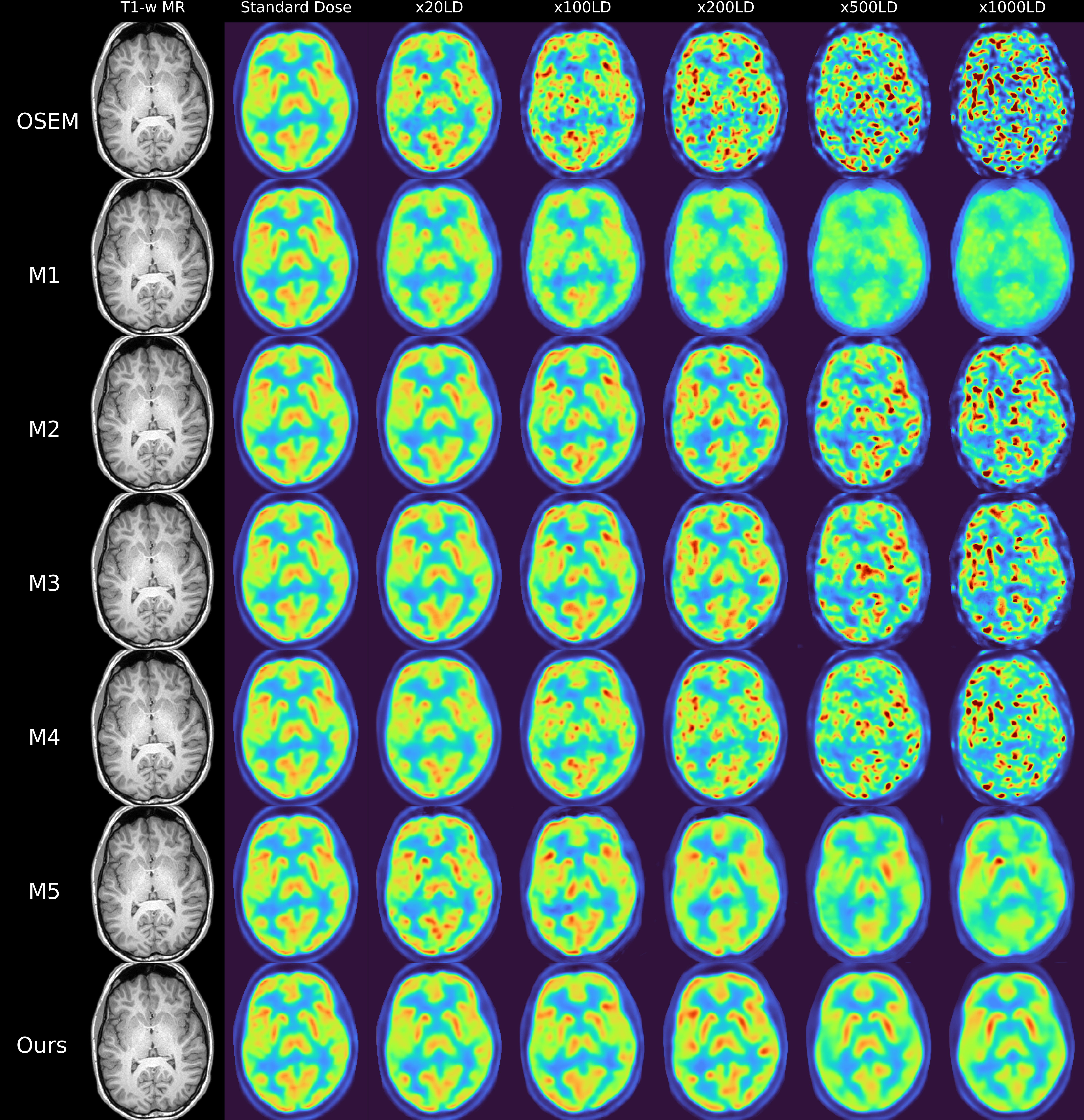}
\caption{Comparison of our method with benchmarks from the literature trained on $\times 20$ low-dose $^{18}F$-FDG images and unpaired MR. The proposed method uses kernel patch sizes of $1,8,32,96,96$ for dose reduction factors $\times 20$,$\times 100$,$\times 200$,$\times 500$,$\times 1000$ respectively. We set $\gamma_{max} = \gamma_{min} = 0$}
\end{figure*}


\begin{figure*}[h!]
\centering
\includegraphics[width=0.95\textwidth]{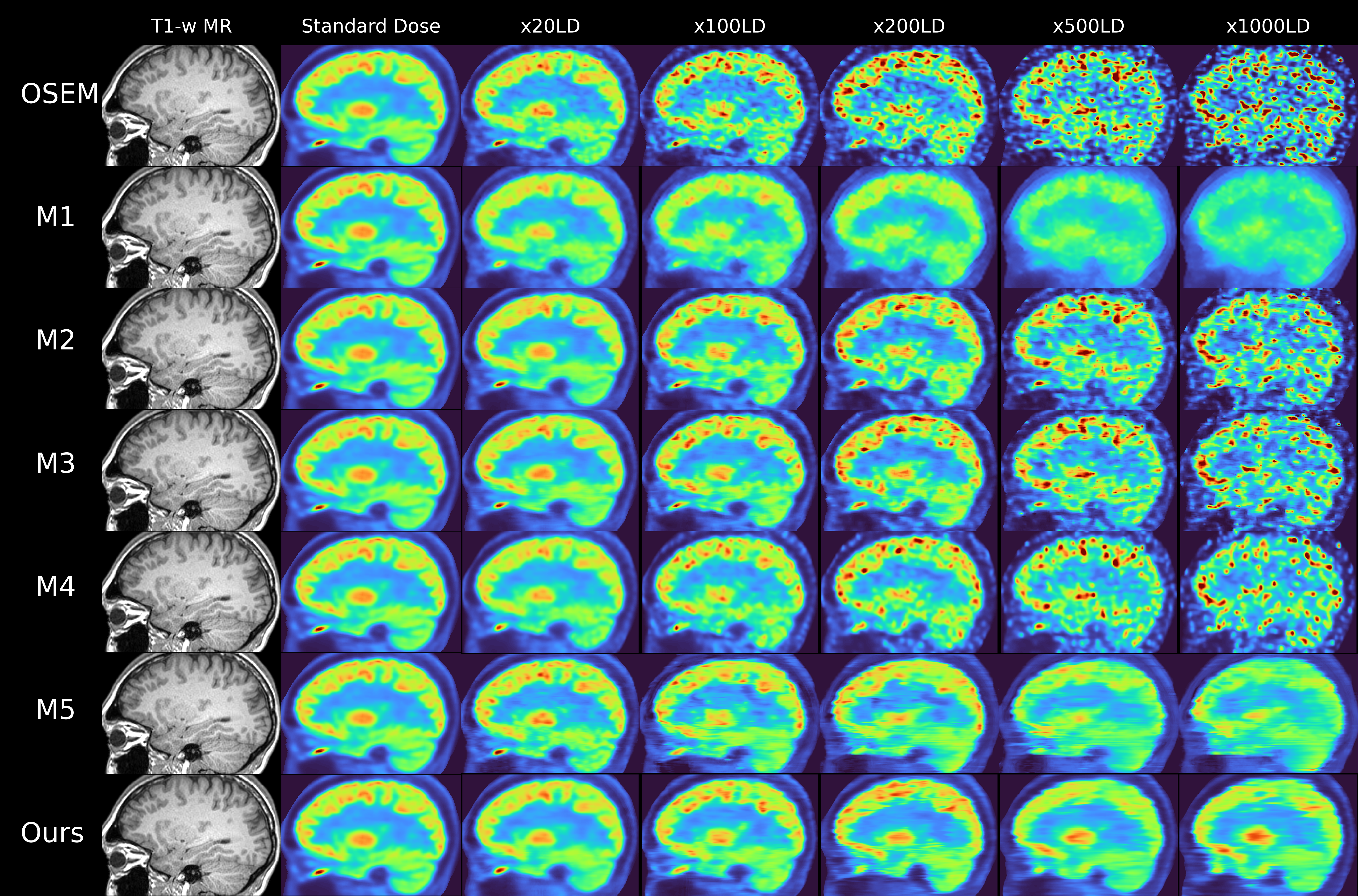}
\caption{\color{black} Comparison of our method with benchmarks from the literature trained on $\times 20$ low-dose $^{18}F$-FDG images and unpaired MR. The proposed method uses kernel patch sizes of $1,8,32,96,96$ for dose reduction factors $\times 20$,$\times 100$,$\times 200$,$\times 500$,$\times 1000$ respectively. We set $\gamma_{max} = \gamma_{min} = 0$}
\end{figure*}

\begin{table}[t!]
\centering
\caption{\footnotesize{Number of trainable parameters, whether methods are re-implemented (re-imp) or source code (source), and supervision method.}}
\begin{tabular}{p{1.1cm} >{\centering\arraybackslash}m{1.6cm} >{\centering\arraybackslash}m{1.6cm}>{\centering\arraybackslash}m{1.8cm}}
\hline
Method & Gen. params & Disc. params &  Implementation \\
\hline
M1   &     0  & 0     & Source   \\
M2   &   1.9M & 0     & re-imp   \\
M3   &   3.9M & 2.4M  & re-imp   \\
M4   &   0.2M & 0     & Source   \\
M5   &   1.9M & 0     & re-imp   \\
Ours &   6.5M & 0.32M &          \\
\hline
\end{tabular}
\vspace{-0.5cm}
\end{table}

\subsection{Reference methods}
 Baseline methods from the literature are for comparison with the proposed method. Table 2 shows the total number of trainable parameters for each method.

 \subsubsection{Bowsher prior (M1)}
 A conventional MR guided PET image reconstruction using a Bowsher prior. We reimplement code as used in \cite{Sudarshan2021_Bowsher}. The hyperparameter controlling the MR prior strength was varied as a function of dose reduction factor. We use a patch size of 9 pixels for each reconstruction. 

 \subsubsection{Multi-modal residual Unet (M2)} 
 A residual Unet consistent with that found in \cite{Chen2019_M1,Chen2021_M1,Ouyang2019_M1} was trained to map T1-w, T2-w and low-dose PET images to a standard-dose PET image. Implementations from the literature included FLAIR contrast MR images and multi-slice inputs. In this work, no FLAIR images were available and 2D inputs were used for consistency with the proposed method. We re-implement the method as described in \cite{Chen2019_M1}.

\subsubsection{Cycle GAN (M3)}  A conditional cycleGAN as found in \cite{Zhou2020_M2, Zhao2020_M2} was re-implemented to map low-dose PET and MR contrast to standard-dose PET using wasserstein distance loss, cycle-consistency loss and L2 reconstruction loss. We use generator networks with architecture consistent of that used in (M2) with a discriminator network as used in \cite{Zhao2020_M2}.

\subsubsection{Unrolled FBSEM (M4)} An instance of the unrolled FBSEM net as from Mehranian and Reader et al. \cite{Mehranian2019_FBSEM} was implemented with a trainable component consisting of a residual connected 6 layer network, with each serial convolutional block consisting of $32$ convolution kernels of size $3\times3$ , batch normalisation and ReLU activation. We implement this using source code provided \cite{Mehranian2019_FBSEM}, with edits made to incorporate the appropriate forward model  
\color{black}sourced from STIR \cite{STIR} and data pre-processing pipeline.We unrolled 12 updates, (3 iterations, 4 OSEM subsets) with unique trainable weights for the 3 iterations. Sinograms were simulated by forward projecting PET image data.\\

\subsubsection{Population level pre-trained DIP(M5)}
\color{black} A population pre-trained deep image prior\cite{Cui2019_DIPImage} was re-implemented using the same backbone architecture as in (M2). We pre-train the network to map T1 and T2 weighted MR images to $\times20$ low-dose PET images. We perform subject-specific fine-tuning at inference. Dose-level specific empirical stopping criteria  \color{black} are defined based on maximising PSNR for testing data. Encoder weights are fixed during fine tuning and only decoder weights are updated.\\

\begin{figure*}[h!]
\centering
\includegraphics[width=0.95\textwidth]{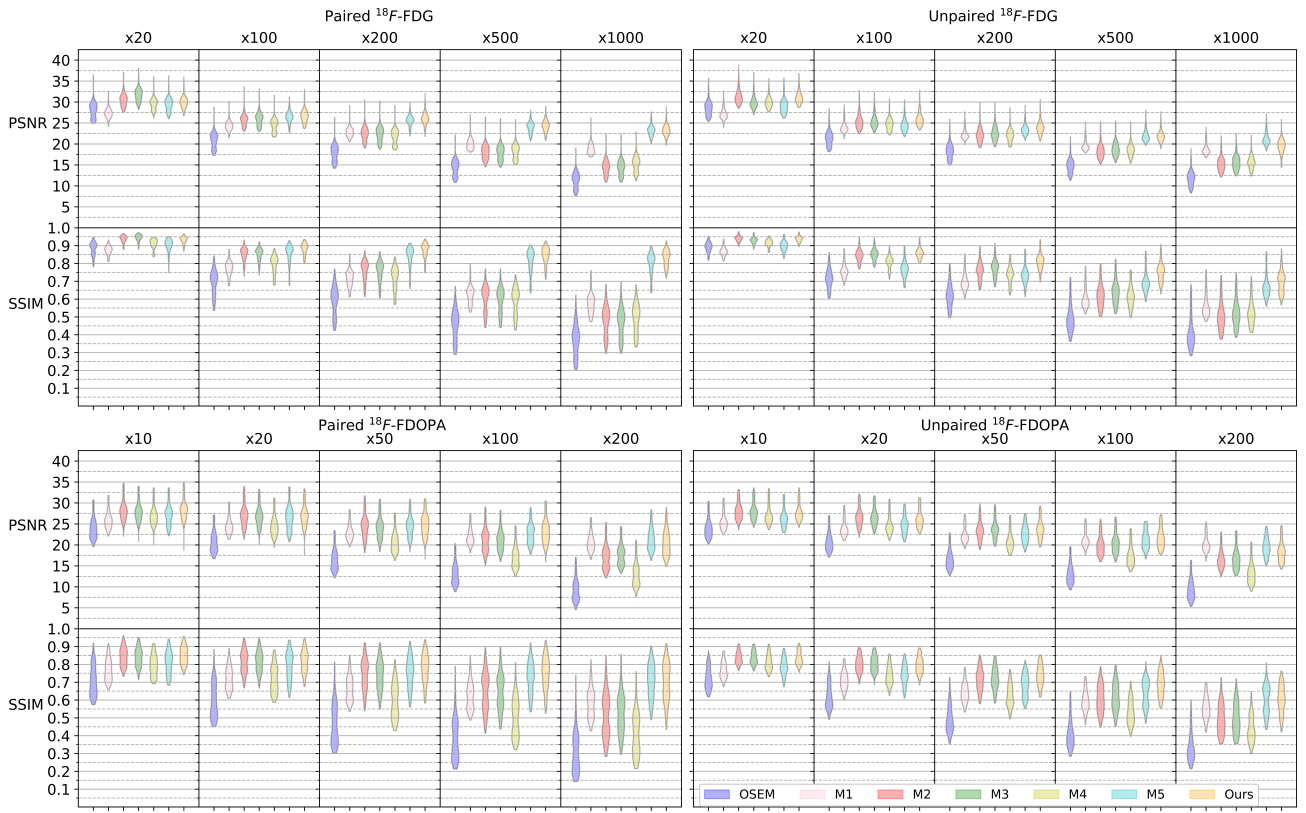}
\caption{Quantitative image quality measures for paired [$^{18}F$]-FDG MR, unpaired [$^{18}F$]-FDG MR, paired [$^{18}F$]-FDOPA MR and unpaired [$^{18}F$]-FDOPA MR.}
\end{figure*}

\vspace{-0.5cm}
\subsection{Training and evaluation of proposed and reference methods}

\color{black} \noindent For paired and unpaired $^{18}F$-FDG datasets, all methods are trained to map $\times 20$ low-dose PET, T1 and T2 weighted MR \color{black}to standard dose PET with performance evaluated on previously unseen $^{18}F$-FDG PET images at previously unseen easily dose
reduction factors of $\times 100$, $\times 200$, $\times 500$ and $\times 1000$. \color{black} For paired and unpaired $^{18}F$-FDOPA datasets, all methods are trained to map $\times 10$ low-dose PET, T1 and T2 weighted MR, to standard dose PET with performance evaluated on previously unseen $^{18}F$-FDOPA PET images with previously unseen dose reduction factors of $\times 20$, $\times 50$, $\times 100$ and $\times 200$. A 5 fold cross validation was performed using two subjects for validation, three for testing and the remaining subjects for training, for paired and unpaired PET-MR datasets. Each subject is comprised of approximately $155\times192\times192$ voxels from which, 5 contiguous axial slices are extracted, providing approximately 3400 inputs for training each cross validation set and approximately 2250 inputs for evaluation across all cross validation sets.

\noindent For each dataset neural networks are trained with patch-sizes ranging from $p=1$ up to $p=96$ where the patch-size is decreased by a factor of two for each subsequent deep layer, maintaining a constant ratio between patch-size and feature map size for each layer. Information maximisation constraints are applied using $l = 0$, and $l=3$ with hyper-parameter strengths $\gamma_{max} \in \{1e^{-1},1e^{-2},1e^{-3} \}$ when kernel-patch size $p >32$. Information minimisation constraints are applied using $l = 0,1,2$, and $l=5$ with hyper-parameter strengths $\gamma_{min} \in \{1e^{-2},1e^{-3},5e^{-4}\}$ when kernel-patch size $p >32$. A stride factor of $s = 1$ is used for kernel patch sizes $p<16$ and a stride factor of $p/8$ for $p = 32,48$. When patch-size is equal to the feature size ($p=96$ for our implementation) only a single patch is extracted and no stride factor is defined.

\subsection{Ablation Study}
We perform an ablation study on several components of our method using the paired $^{18}F$-FDG MR-PET dataset.
\subsubsection{Latent-space kernel functions}

Performance is evaluated as a function of patch size as defined in equation (12). We compare linear latent-space kernel functions defined as in equation(13) with Gaussian radial basis functions (RBF) defined element-wise as
\begin{equation}
k^{l,c}_{i,j} = \begin{cases} 
      exp(-\tfrac{1}{2\sigma} ||\mathbf{b}^{l,c}_i - \mathbf{b}^{l,c}_j||^2) &  j \in \mathcal{N}_{p}(i)\\
      0 & \text{elsewhere}
      \end{cases}
\end{equation}
where $\mathbf{b}^{l,c}_i$ is defined in equation (11) and $\sigma = 0.01$. We compare linear and RBF kernels in terms of PSNR and SSIM for varying patch-sizes. Images of kernel features and rows \color{black} from latent-space kernel functions are presented to highlight the function of latent-space kernel functions in providing MR guidance.

\subsubsection{Information Constraints}
We evaluate the effect of varing $\gamma_{max}$ and $\gamma_{min}$in terms of PSNR and SSIM across all dose reduction factors. We evaluate information maximisation \color{black} constraints using patches from $l=0$ and deep features varying from $l=3,4,5$. We evaluation information minimisation constraints using combinations of shallow kernel features $l=\{0\}$,$l=\{0,1\}$ and $l=\{0,1,2\}$, and deep PET features from $l=5$.

\begin{figure*}[h!]
\centering
\includegraphics[width=0.95\textwidth]{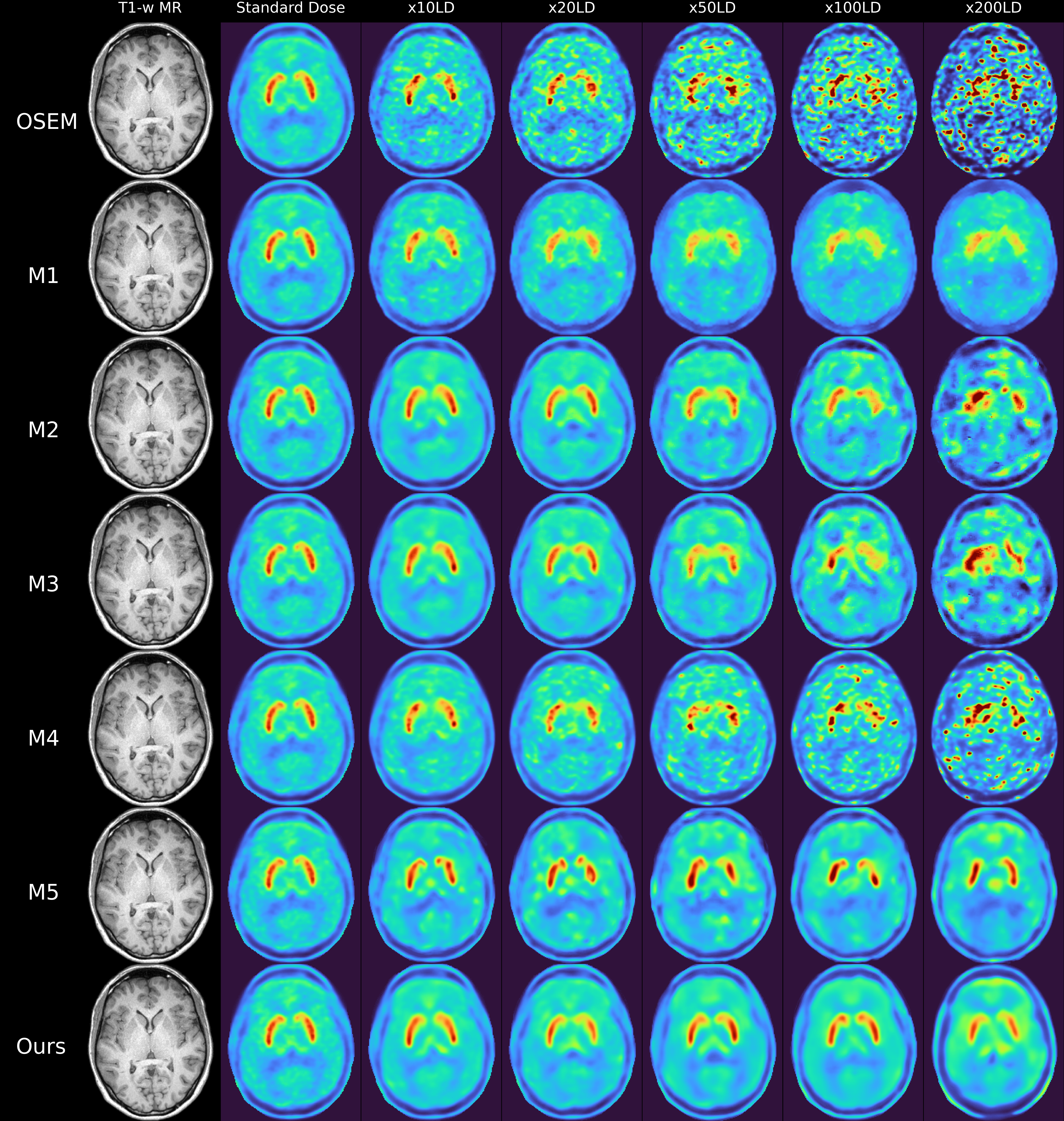}
\caption{Comparison of our method with benchmarks from the literature trained on $\times 10$ low-dose $^{18}F$-FDOPA images and unpaired MR. The proposed method uses kernel patch sizes of $4,4,32,48,96$ for dose reduction factors $\times 10$,$\times 20$,$\times 50$,$\times 100$,$\times 200$ respectively. We apply information constraints $\gamma_{max} = 0.005$ and $\gamma_{min} = 0.001$ for dose reduction factors of $\times 100$ and $\times 200$.}

\end{figure*}


\begin{figure*}[h!]
\centering
\includegraphics[width=0.95\textwidth]{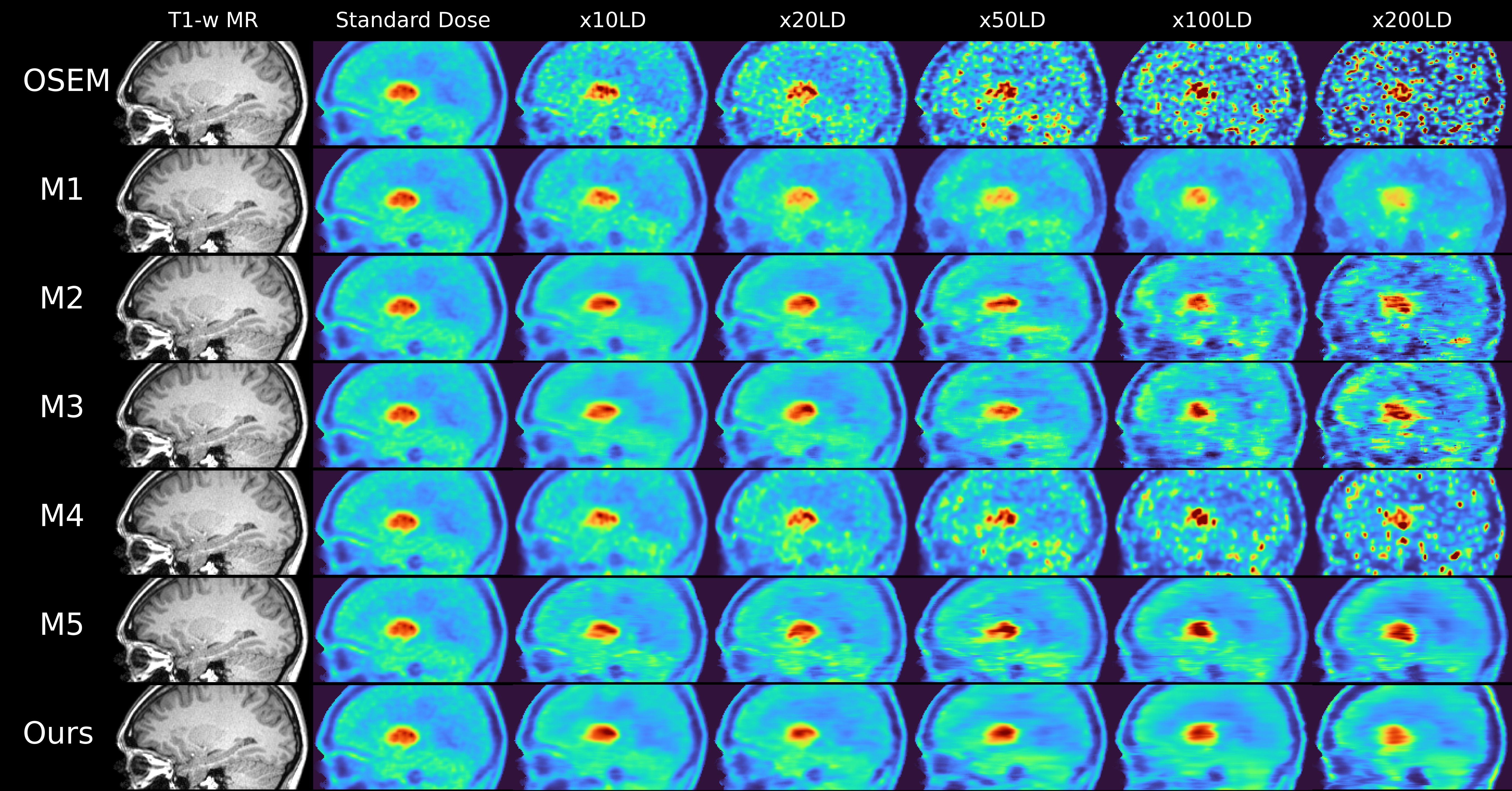}
\caption{Comparison of our method with benchmarks from the literature trained on $\times 10$ low-dose $^{18}F$-FDOPA images and unpaired MR. The proposed method uses kernel patch sizes of $4,4,32,48,96$ for dose reduction factors $\times 10$,$\times 20$,$\times 50$,$\times 100$,$\times 200$ respectively. We apply information constraints $\gamma_{max} = 0.005$ and $\gamma_{min} = 0.001$ for dose reduction factors of $\times 100$ and $\times 200$.}
\end{figure*}

\section{Results}
\color{subsectioncolor}\subsection{Paired MR and $^{18}F$-FDG brains}
\color{black}\noindent The proposed method preserves structural features considerably better than the benchmark methods when applied to unseen dose reduction images containing minimal anatomical information as is demonstrated in figure 2 and figure 3. Baseline methods (M2) to (M5) provide excellent in-distribution performance, yet generalise  poorly across previously unseen reduced dose images. Methods (M1) and (M5) provide better results for large dose reduction factors relative to baseline supervised methods, however demonstrate poorer performance on $\times 20$ low-dose. Variations in performance between supervised methods may be attributed to the number of parameters used for each method as listed in table 2. As (M5) is an unrolled reconstruction algorithm with explicit data-consistency steps and considerably smaller number of trainable parameters, it is expected for further low-dose data noise artefacts will be more apparent. Similary, methods (M2) and (M3) which learn much deeper latent-space representations are likely to provide more unpredictable results in out-of-distribution cases. Our proposed method shows considerably improved contrast recovery as demonstrated in table 3. While (M5) also performs well for very low-dose data, the proposed method performs considerably better recovering contrast in the caudate and putamen, as signal from small anatomical regions is lost at large dose-reduction factors and unretrievable using self-supervised methods. Figure 8 shows performance in terms of PSNR and SSIM for each method.

\begin{table*}[t!]
\centering
\caption{\footnotesize{Mean normalised bias and variance in regional uptake for paired and unpaired $^{18}F$-FDG and MR images. Mean normalised regional variance is shown in parenthesis. Values are expressed as a percentage.}}
\begin{tabular}{p{0.1cm} p{1.00cm}  p{1.16cm} p{1.16cm} p{1.16cm} p{1.16cm} p{1.16cm} | p{1.16cm} p{1.16cm} p{1.16cm} p{1.16cm} p{1.16cm}}
\hline & \multicolumn{6}{c|}{Paired $^{18}F$-FDG Dose Reduction Factor} &  \multicolumn{5}{c}{Unpaired $^{18}F$-FDG Dose Reduction Factor}\\
 & & $\times 20$ &  $\times 100$ & $\times 200$  & $\times 500$ & $\times 1000$ &$\times 20$ &  $\times 100$ & $\times 200$  & $\times 500$ & $\times 1000$\\
\hline
                                              & OSEM     &  7.3(1.0)           & 16(4.9)              & 23(9.9)             & 35(21)               & 51(46)           & 7.1(0.9)           & 16(5.0)           & 23(10)              & 35(22)            & 48(43) \\
\multirow{7}{0.5cm}{\rotatebox{90}{\  \  \ \ \  Gray Matter}} & M1       &  9.0(1.2)           & 12(2.0)              & 14(2.5)             & 22(4.2)              & 25(4.8)          & 9.0(1.2)           & 13(2.4)           & 15(3.4)             & 23(5.7)           & 26(6.7)\\ 
                                              & M2       &  5.3(0.5)           & 8.8(1.4)             & 13(3.2)             & 27(12)               & 35(21)           & \textbf{5.0(0.5)}  & 9.8(1.8)          & 15(3.9)             & 24(9.1)           & 34(19) \\ 
                                              & M3       &  \textbf{4.8(0.4)}  & 8.7(1.4)             & 13(3.2)             & 23(9.1)              & 38(24)           & 5.8(0.6)           & 9.7(1.7)          & 14(3.6)             & 22(8.1)           & 33(17) \\
                                              & M4       &  6.1(0.6)           & 11(2.1)              & 15(4.1)             & 23(8.6)              & 35(20)           & 5.9(0.6)           & 11(2.1)           & 15(4.1)             & 23(8.8)           & 32(18) \\ 
                                              & M5       &  5.9(0.6)           & \textbf{7.7(1.0)}    & 8.4(1.2)            & \textbf{11(1.8)}              & 13(2.2)          & 6.6(0.8)           & 11(1.9)           & 12(2.2)             & 14(2.9)           & 16(3.4)\\ 
                                              & Ours     &  5.6(0.5)           & 7.9(1.1)             & \textbf{8.0(1.1)}   & \textbf{11(1.7)}     & \textbf{11(2.1)} & 5.1(0.5)           & \textbf{8.7(1.4)} & \textbf{11(2.2)}    & \textbf{12(2.5)}  & \textbf{13(3.0)}\\
 \hline
                                            & OSEM & 8.2(1.2)          & 19(6.3)          & 26(13)            & 40(31)            & 57(69)           & 8.0(1.1)          & 19(6.3)         & 27(13)           & 40(30)          & 56(67)  \\
\multirow{7}{0.5cm}{\rotatebox{90}{\  \  \ \ \  White matter }} & M1   & 9.3(1.4)          & 12(2.1)          & 13(2.5)           & 16(3.8)           & 18(4.9)          & 10(1.1)           & 14(2.1)         & 16(2.8)          & 19(4.8)         & 20(5.7) \\ 
                                            & M2   & 5.6(0.5)          & 9.7(1.6)         & 14(3.4)           & 27(13)            & 38(30)           & \textbf{5. 5(0.6)} & 11(2.3)         & 16(4.8)          & 26(12)          & 38(27)  \\ 
                                            & M3   & \textbf{5.0(0.4)} & 9.9(1.6)         & 14(3.6)           & 25(13)            & 41(33)           & 6.1(0.7)          & 11(2.1)         & 15(4.1)          & 24(10)          & 35(23)  \\
                                            & M4   & 7.4(0.7)          & 12(2.2)          & 14(3.8)           & 22(9.5)           & 32(22)           & 7.4(0.7)          & 11(2.1)         & 15(4.0)          & 22(9.7)         & 31(21)  \\ 
                                            & M5   & 6.7(0.8)          & 9.0(1.4)         & 9.7(1.6)          & 11(2.1)           & 13(2.3)          & 7.3(0.9)          & 13(2.6)         & 15(3.4)          & 16(4.0)         & 17(4.3) \\ 
                                            & Ours & 5.9(0.6)          & \textbf{8.7(1.3)} & \textbf{8.1(1.2)} &\textbf{ 9.8(1.7)} & \textbf{11(2.2)} & 5.6(0.6)          & \textbf{10(1.9)} & \textbf{13(3.0)} & \textbf{14(3.5)} & \textbf{15(3.9)} \\
 \hline
                                           & OSEM & 7.4(0.9)          & 19(6.0)          & 25(11)            & 38(26)             & 57(63)                & 7.4(0.9)           & 19(6.1)          & 25(10)            & 38(26)            & 52(54) \\
\multirow{7}{0.5cm}{\rotatebox{90}{\  \  \ \  Thalamus\ \ }} & M1   & 8.6(0.8)          & 11(1.4)          & 14(2.0)           & 18(3.2)            & 19(3.7)               & 6.7(0.7)           & 10(1.5)          & 13(2.1)           & 19(4.4)           & 20(4.6)  \\ 
                                           & M2   & \textbf{4.4(0.3)} & 9.0(1.3)        & 13(2.8)          & 31(10)            & 36(23)                & \textbf{4.4(0.3)}  & 12(2.3)          & 16(4.2)           & 25(10)            & 35(21)  \\ 
                                           & M3   & \textbf{4.4(0.3)} & 9.7(1.6)        & 13(3.0)          & 24(9.9)           & 41(29)                & 4.5(0.3)           & 9.0(1.3)         & 13(2.7)           & 22(8.0)           & 32(17)  \\
                                           & M4   & 6.3(0.6)          & 11(2.1)          & 15(3.4)           & 23(9.0)            & 36(23)                & 6.3(0.6)           & 12(2.2)          & 15(3.5)           & 23(9.0)           & 30(19)\\ 
                                           & M5   & 5.9(0.6)          & \textbf{7.9(1.1)}& 7.4(0.9)         & 11(1.4)           & \textbf{10(1.6)}    & 6.5(0.7)           & 9.7(1.5)         & 10(1.1)           & 11(1.9)           & 12(2.1)  \\ 
                                           & Ours & 4.8(0.4)          & 8.0(1.1)         & \textbf{6.4(0.6)} & \textbf{9.0(1.1)} & \textbf{10(1.3)}     & 4.4(0.3)           & \textbf{7.5(0.9)}& \textbf{9.2(1.3)} & \textbf{7.9(1.0)} & \textbf{9.8(1.4)}\\
 \hline
                                          & OSEM         & 7.5(1.0)         & 20(7.0)          & 25(11)          & 40(29)          & 50(36)           & 7.7(1.1)          & 20(6.6)          & 25(11)          & 39(27)          & 51(51)  \\
\multirow{7}{0.5cm}{\rotatebox{90}{\  \ \ \ \ \  Caudate\ \ }} & M1           & 10(1.4)          & 14(2.4)          & 18(3.3)         & 25(5.3)         & 26(5.8)          & 13(2.4)           & 16(3.2)          & 19(3.8)         & 25(5.8)         & 26(6.5)\\ 
                                          & M2           & 5.6(0.5)         & 10(1.7)          & 15(3.4)         & 33(12)          & 38(25)           & \textbf{5.3(0.5)} & 12(2.5)          & 16(4.3)         & 29(12)          & 36(22) \\ 
                                          & M3           & \textbf{5.1(0.4)}& 11(2.2)          & 14(3.1)         & 27(12)          & 39(17)           & 5.7(0.6)          & 9.9(1.7)         & 13(2.7)         & 27(9.7)         & 32(15) \\
                                          & M4           & 9.4(1.2)         & 15(3.1)          & 16(3.9)         & 28(12)          & 37(16)           & 9.0(1.0)          & 14(3.0)          & 16(4.2)         & 28(12)          & 34(21) \\ 
                                          & M5           & 6.5(0.7)         & 10(1.6)          & 11(1.9)         & 16(2.7)         & 14(2.3)          & 7.2(0.9)          & 15(3.9)          & 14(3.3)         & 19(4.8)         & 20(5.5) \\ 
                                          & Ours         & 6.6(0.9)         & \textbf{9.6(1.6)}& \textbf{10(1.3)}& \textbf{11(1.9)}& \textbf{12(2.0)} & 5.6(0.5)          & \textbf{9.1(1.4)}& \textbf{11(1.6)}& \textbf{12(2.3)}& \textbf{15(3.3)}  \\
 \hline
                                          & OSEM & 7.0(0.8)          & 17(4.7)           & 24(10)          & 35(22)               & 49(40)          & 7.0(0.8)         & 17(4.7)           & 24(10)            & 35(22)           & 50(46)   \\
\multirow{7}{0.5cm}{\rotatebox{90}{\  \  \ \ \ \  Putamen\ \ }} & M1   & 9.1(0.9)          & 11(1.3)           & 13(1.4)         & 20(2.5)              & 23(3.2)         & 7.3(0.7)         & 11(1.5)           & 12(1.7)           & 21(2.6)          & 23(3.1)  \\ 
                                          & M2   & 4.9(0.4)          & 8.6(1.2)          & 12(2.5)         & 28(10)               & 34(20)          & 4.7(0.4)         & 9.7(1.5)          & 14(3.6)           & 24(8.7)          & 35(19)   \\ 
                                          & M3   & 4.9(0.3)          & 8.7(1.2)          & 12(2.7)         & 22(7.6)              & 36(20)          & \textbf{4.7(0.3)}& 8.0(1.0)          & 12(2.3)           & 21(6.4)          & 32(16)   \\
                                          & M4   & 6.0(0.5)          & 10(1.7)           & 13(3.2)         & 22(7.4)              & 33(15)          & 6.0(0.5)         & 10(1.7)           & 14(3.3)           & 22(7.7)          & 31(17)   \\ 
                                          & M5   & 5.9(0.6)          & 8.2(1.1)          & 9.4(1.4)        & 15(1.9)              & 16(2.3)         & 6.1(0.6)         & 10(1.5)           & 10(1.6)           & 13(2.2)          & 15(3.0)  \\ 
                                          & Ours & \textbf{4.8(0.4)}  & \textbf{7.7(1.0)}  & \textbf{8.3(1.0)}& \textbf{8.9(1.2)} & \textbf{11(1.3)}& 5.0(0.3)         & \textbf{7.7(1.0)} & \textbf{9.2(1.3)} & \textbf{8.9(1.2)}& \textbf{9.4(1.4)} \\
 \hline 
\end{tabular}
\end{table*}

\subsection{Paired MR and $^{18}F$-FDOPA  brains}
\noindent Anatomical consistency is well preserved using the proposed method for all dose reduction factors as shown in figure 4 and figure 5, with the good contrast between midbrain features and the white matter of particular note. For dose reduction factors up to $\times 200$, the proposed method recovers contrast of brain regions better than baseline methods as shown in table 4. Supervised learning methods (M2) to (M4) generalise comparatively poorly across further reduced dose images. Baseline methods (M1) and (M5) maintain anatomical consistency, however both demonstrate biasing at large dose reduction factors. Figure 8 shows the PSNR and SSIM for each method.

\subsection{Unpaired MR and $^{18}F$-FDG brains}
\noindent Figure 6 and figure 7 compares the proposed and reference methods using low-dose $^{18}F$-FDG PET and unpaired MR. Similar to paired $^{18}F$-FDG and MR, structural features are preserved well across all dose reduction factors. The average performance of all methods relative to paired PET MR data is decreased, however this strongly depends on the quality of coregistration to the template MR on an individual basis. Table 3 shows bias and variance measures for varying dose reduction. We see that our method is still able to utilise the template MR to achieve good contrast recovery, particularly for small anatomical regions which are sensitive to reduced dose. As the template MR provides lower fidelity anatomical information, explicitly constraining latent representations using a population level kernel function reduces performance. Performance in terms of PSNR and SSIM is shown in figure 8.

\begin{table*}[t!]
\centering
\caption{\footnotesize{Mean normalised bias and variance in regional uptake for paired and unpaired $^{18}F$-FDOPA and MR images. Mean normalised regional variance is shown in parenthesis. Values are expressed as a percentage.}}
\begin{tabular}{p{0.1cm} p{1.00cm}  p{1.16cm} p{1.16cm} p{1.16cm} p{1.16cm} p{1.16cm} | p{1.16cm} p{1.16cm} p{1.16cm} p{1.16cm} p{1.16cm}}
\hline & \multicolumn{6}{c|}{Paired $^{18}F$-FDOPA Dose Reduction Factor} &  \multicolumn{5}{c}{Unpaired $^{18}F$-FDOPA Dose Reduction Factor}\\
&  & $\times 10$ &  $\times 20$ & $\times 50$  & $\times 100$ & $\times 200$ &$\times 10$ &  $\times 20$ & $\times 50$  & $\times 100$ & $\times 200$\\
\hline
                                              & OSEM     & 12  (2.6)         & 18(5.3)           &28(14)             & 41(30)            & 59 (73)             & 12(2.4)           & 17(5.1)           & 28(14)            & 42(31)          & 59(74) \\
\multirow{7}{0.5cm}{\rotatebox{90}{\  \  \ \ \  Gray matter\ \ }} & M1       & 9.7 (1.7)         & 11(1.9)           & 12(2.4)           & 15(4.0)           & 17 (4.6)            & 13(2.2)           & 13(2.5 )          & 14 (2.5)          & 14(2.6)         & 15(3.2)   \\ 
                                              & M2       & \textbf{6.8 (0.8)}& 7.6(1.0)          & 9.3(1.5)          & 14(3.4)           & 23 (9.0)            &  \textbf{6.4(0.7)} &  \textbf{7.5(1.0)} & 11 (2.0)          & 17(4.6)         & 26(11)  \\ 
                                              & M3       & \textbf{6.8 (0.8)}&  \textbf{7.5(1.0)}& 10(1.8)           & 14(3.0)           & 20 (6.2)            & 6.7(0.8)          &  \textbf{7.5(1.0)} & 11 (1.9)          & 16(4.2)         & 25(10)   \\
                                              & M4       & 7.8 (1.1)         & 11(2.0)           & 16(4.3)           & 23(9.7)           & 35 (25)             & 7.4(0.9)          & 10(1.8)           & 17 (4.8)          & 25(11)          & 38(30)   \\ 
                                              & M5       & 7.4 (1.0)         & 8.2(1.2)          & 8.8(1.3)          & 11(1.9)           & 13 (2.3)            & 8.3(1.2)          & 9.2(1.4)          & 11 (2.0)          &  \textbf{12(2.3)} & 19(6.5)  \\ 
                                              & Ours     & 7.0 (0.9)         & 8.4(1.2)          &  \textbf{8.5(1.3)}&  \textbf{9.5(1.6)}& \textbf{9.2(1.4)}   & 6.5(0.7)          & 7.6(1.0)          & \textbf{8.7 (1.3)}&  \textbf{12(2.5)} &  \textbf{13(2.8)}   \\
 \hline
                                               & OSEM     & 13(2.7 )        & 18(5.7)          & 30(16)            & 42(32)           & 60(75)          & 12.6(2.7)        & 18(5.7)          & 29(15)           & 43(35)          & 60(78) \\
\multirow{7}{0.5cm}{\rotatebox{90}{\  \  \ \ \  White matter }} & M1       & 8.4(1.3)        & 9.2(1.5)         & 11(2.0)           & 13(3.2)          & 16(4.3)         & 13(1.9)          & 14(2.2)          & 15(2.8)          & 15(2.7)         & 17(4.2) \\ 
                                               & M2       & 6.6(0.8)        & 7.5(1.0)         & 9.1(1.5)          & 14(3.6)          & 23(9.7)         & \textbf{6.6(0.7)}& 7.6(1.0)         & 11(2.1)          & 17(4.8)         & 26(11)   \\ 
                                               & M3       & \textbf{6.2(0.7)}& \textbf{6.9(0.8)}& 9.3(1.6)          & 13(3.1)          & 21(8.0)         & 6.8(0.8)         & 7.7(1.0)         & 11(2.0)          & 17(4.6)         & 27(12)   \\
                                               & M4       & 7.3(0.9)        & 10(1.8)          & 16(4.5)           & 23(9.5)          & 34.0(24)        & 7.4(0.9)         & 10(1.8)          & 16(4.8)          & 24(11)          & 37(29)\\ 
                                               & M5       & 7.3(1.0)        & 7.7(1.1)         & 8.8(1.5)          & 11(2.1)          & 12(2.8)         & 8.4(1.3)         & 9.3(1.6)         & 11(2.3)          & \textbf{12(2.5)}& 17(5.9)\\ 
                                               & Ours     & 6.4(0.8)        & 8.4(1.3)         & \textbf{8.1(1.3)} & \textbf{9.3(1.5)}&\textbf{9.2(1.6)}& \textbf{6.6(0.7)}& \textbf{7.5(1.0)}&\textbf{8.8(1.4)} & \textbf{12(2.7)}& \textbf{12(2.9)} \\
 \hline   
                                           & OSEM & 13  (2.8)         & 19(5.7)          & 29(13)          & 43(29)           & 58(69)           &  13(2.6)          &19(6.1)          & 29(14)         & 44(33)         & 61(73)  \\
\multirow{7}{0.5cm}{\rotatebox{90}{\  \  \ \ \  Thalamus\ \ }} & M1   & 9.9 (1.3)         & 11(1.7)          & 12(2.0)         & 17(2.7)          & 23(4.2)          &  15(1.6)          &17(2.0)          & 17(2.3)        & 16(3.2)        & 21(5.0) \\ 
                                           & M2   & 7.4 (0.8)         & \textbf{8.7(1.1)}& 12(1.9)         & 17(3.8)          & 29(11)           &  8.1(1.0)         &\textbf{9.6(1.5)}&13(2.2)         & 21(4.5)        & 30(9.2) \\ 
                                           & M3   & 8.0 (1.0)         & 9.2(1.3)         & 11(1.8)         & 18(4.1)          & 26(9.4)          &  7.9(1.0)         &9.9(1.5)         &13(2.4)         & 19(5.0)        & 35(12)  \\
                                           & M4   & 9.1 (1.1)         & 12(2.1)          & 16(3.7)         & 24(7.8)          & 35(24)           &  8.8(1.0)         &12(2.2)          &17(4.0)         & 25(8.6)        & 40(26)  \\ 
                                           & M5   & 9.9 (1.6)         & 11(2.1)          & 15(2.8)         & 20(5.3)          & 20(3.9)          &  10(1.5)          &13(2.6)          &14(3.1)         & 18(3.4)        & 23(6.4) \\ 
                                           & Ours & \textbf{6.9 (0.7)}& 11(1.9)          & \textbf{10(1.4)}& \textbf{9.9(1.2)}& \textbf{9.9(1.3)}& \textbf{7.8(0.9)} &9.7(1.4)         &\textbf{10(1.5)}& \textbf{15(2.5)}&\textbf{19(5.2)} \\
 \hline
                                          & OSEM & 12(2.1)         & 17(5.0)          & 26(13)         & 38(26)          & 55(70)           &12(27)           & 17(5.3)          & 29(16)         & 39(27)         & 61(78)  \\
\multirow{7}{0.5cm}{\rotatebox{90}{\  \  \ \ \  Caudate\ \ }} & M1   & 13(2.7 )        & 14(3.2)          & 17(4.4)        & 23(5.3)         & 25(6.2)          &17(4.2)          & 18(4.4)          & 19(6.2)        & 19(5.8)        & \textbf{24(9.2)}  \\ 
                                          & M2   & 7.9(1.0)        & 9.8(1.7)         & 13(2.7)        & 19(4.8)         & 24(8.7)          &\textbf{8.8(1.3)}& 9.8(1.6)         & 14(3.3)        & 20(4.3)        & 33(10)  \\ 
                                          & M3   & 7.8(0.8)        & \textbf{8.6(1.1)}&\textbf{11(2.0)}& \textbf{16(3.4)}& 26(8.2)          &9.9(1.4)         & \textbf{9.7(1.5)}& 14(3.4)        & 18(4.4)        & 29(11)  \\
                                          & M4   & 11(1.5)         & 12(2.0)          & 16(4.3)        & 27(9.7)         & 34(23)           &9.7(1.4)         & 12(2.4)          & 18(6.1)        & 25(10)         & 37(30)  \\ 
                                          & M5   & 10(1.6)         & 13(2.5)          & 15(3.8)        & \textbf{16(4.2)}& 19(6.0)          &12(2.2)          & 12(2.3)          & 15(3.9)        & 20(6.0)        & 29(12)\\ 
                                          & Ours &\textbf{7.7(1.0)}& 12(2.6)          & 12(2.1)        & 17(3.4)         & \textbf{14(2.3)} &10(1.4)          & 11(1.9)          & \textbf{13(2.5)}& \textbf{14(2.8)}& \textbf{24(8.8)}\\
 \hline
                                          & OSEM & 11(1.8)          & 15(3.4)          & 26(11)          & 34(20)         & 47(44)            & 12(2.1)          & 15(3.6)         & 28(13)          & 40(18)          & 52(51)  \\
\multirow{7}{0.5cm}{\rotatebox{90}{\  \  \ \ \  Putamen\ \ }} & M1   & 10(1.2)          & 12(1.1)          & 13(1.8)         & 25(2.7)        & 28(3.0)           & 19(2.9)          & 18(3.5)         & 18(4.0)         & 17(3.9)         & 21(4.9) \\ 
                                          & M2   & 7.3(0.9)         & 7.9(1.0)         & \textbf{11(2.0)}& 16(3.6)        & 23(7.9)           & \textbf{7.1(0.7)}& 8.0(1.0)        & 13(2.2)         & 17(3.7)         & 23(8.4) \\ 
                                          & M3   & \textbf{6.5(0.7)}& \textbf{6.9(0.7)}& 12(2.2)         & 16(4.3)        & 23(8.0)           & 8.5(1.0)         & \textbf{7.6(0.9)}& 13(2.3)         & 19(4.2)         & 24(9.0) \\
                                          & M4   & 8.0(0.9)         & 9.6(1.4)         & 16(4.1)         & 22(7.8)        & 29(15)            & 8.1(1.0)         & 10(1.6)         & 17(4.9)         & 24(9.0)         & 33(21)  \\ 
                                          & M5   & 8.1(1.1)         & 9.0(1.3)         & 13(2.8)         & 15(3.6)        & 18(4.5)           & 9.9(1.4)         & 10(1.7)         & 15(3.5)         & 16(4.1)         & 29(7.2) \\ 
                                          & Ours & 8.0(1.1)         & 8.8(1.2)         & 12(2.0)         & \textbf{11(1.6)}& \textbf{14(2.9)}   & 7.4(0.8)         & 8.8(1.3)        & \textbf{11(1.3)} & \textbf{12(2.0)}& \textbf{16(3.3)}\\
 \hline 
\end{tabular}
\end{table*}

\subsection{Unpaired MR and $^{18}F$-FDOPA brains}
\noindent Anatomical features are well preserved in the proposed method relative to reference methods as shown in figure 9 and figure 10, and also compare well with results from paired $^{18}F$-FDOPA PET-MR. For dose reduction factor of $\times 200$, we see discrepancies between the predicted standard-dose image and the true standard dose image in the posterior region of the brain due to mismatching between the MR template and the PET data. Despite this, our method performs considerably better than baseline methods for contrast recovery as demonstrated in table 4. While (M4) performs the best out of baseline methods, it still demonstrates substantial bias in the midbrain region. Sagittal views for template MR data begin to show the limitation of the 2.5D approach for (M2) to (M5), where discrepancies between axial views show small discontinuities.  Figure 8 shows PSNR and SSIM for each method.


\subsection{Ablation Study}
\subsubsection{Latent-space kernel functions}
Figure 11 shows the performance of the proposed method for paired $^{18}F$-FDG brains, using varying patch-sizes in the latent-space kernel functions for an RBF kernel function as shown in equation (24), and the linear kernel function shown in equation (13). Increasing patch-size provides more MR guidance and increases generalisability across the spectrum of dose reduction factors. The linear and Gaussian RBF show comparable performance. For a given patch-size, the generalisability and peak-performance trade-off varies between linear and Gaussian RBF kernels, although equivalent performance appears achievable with each method by varying the kernel patch size.

Figure 12 shows visualisations of an out-of-distribution $\times 1000$ low-dose latent-space code vector, rows from a latent-space kernel matrix and the resultant kernel features for paired $^{18}F$-FDG MR-PET brains. For small patch-sizes, latent-space kernel functions are more sparse, applying less MR guidance and becoming more more anatomically consistent with PET code vectors $\alpha_{\lambda}^{0}$. Increasing the patch size forces the network to utilise MR information for anatomical consistency. An increased kernel stride factor reduces the computational burden of latent-space kernel matrices, however limits the PET information encoded into kernel features. Experiments for patch size $p=32$ and stride factors $s \in \{4,32\}$ for paired $^{18}F$-FDG MR, produce (PSNR,SSIM) values of $(28.1,0.91)$ and $(27.7,0.89)$ for $\times 20$ low-dose images respectively and $(22.9,0.84)$ and $(24.1,0.86)$ for $\times 1000$ low-dose images respectively. While a large stride factor appears to provide only small compromise to in-distribution modelling, this will be highly dependent on the nature of the data. In this case, the anatomical information in brain MR is highly consistent with $^{18}F$-FDG PET.

\begin{figure}[t!]
\centering
\includegraphics[width=0.49\textwidth]{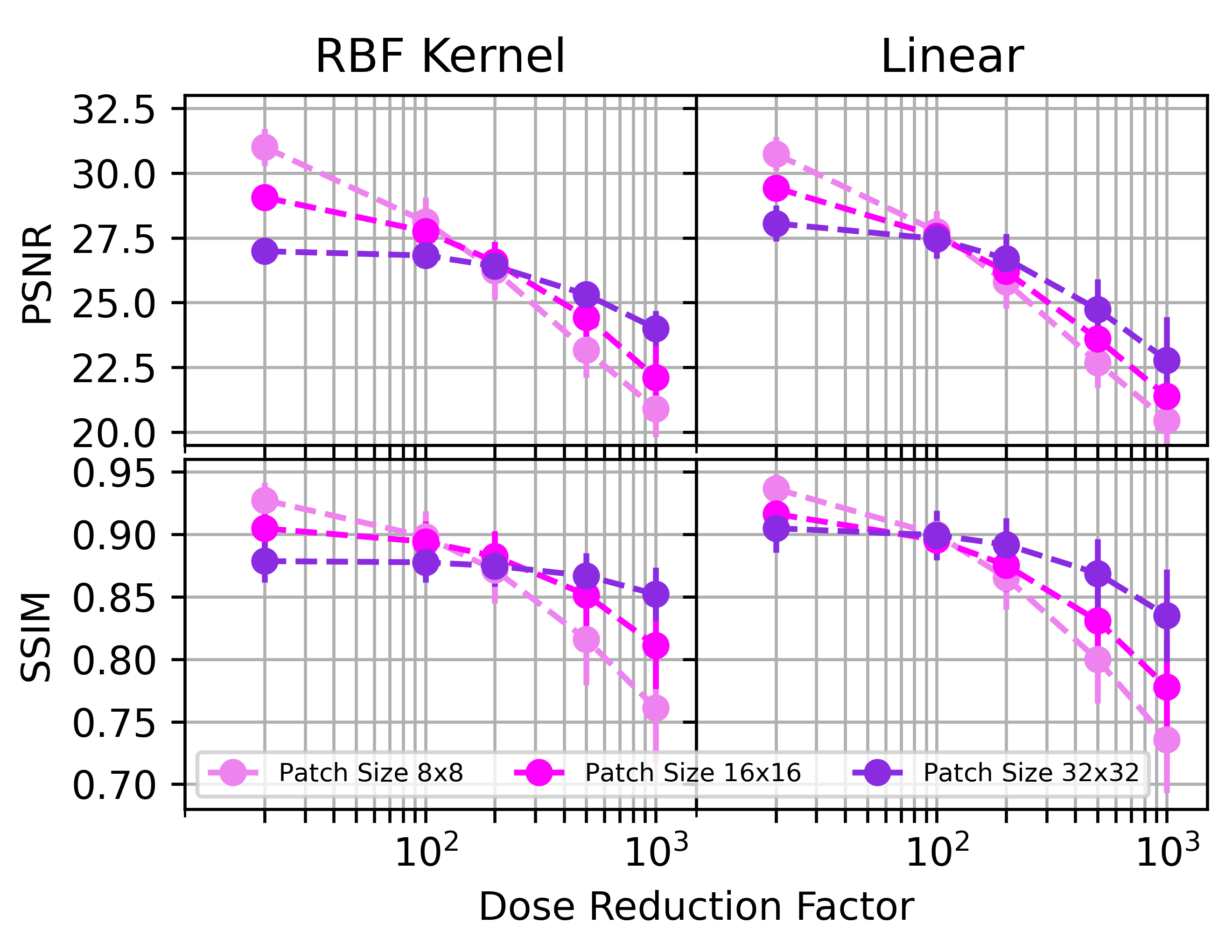}
\caption{Performance of the proposed method as a function of kernel neighbourhood size for linear and RBF kernel functions. Performance is evaluated with no information constraints.}
\end{figure}

\begin{figure}[t!]
\centering
\includegraphics[width=0.49\textwidth]{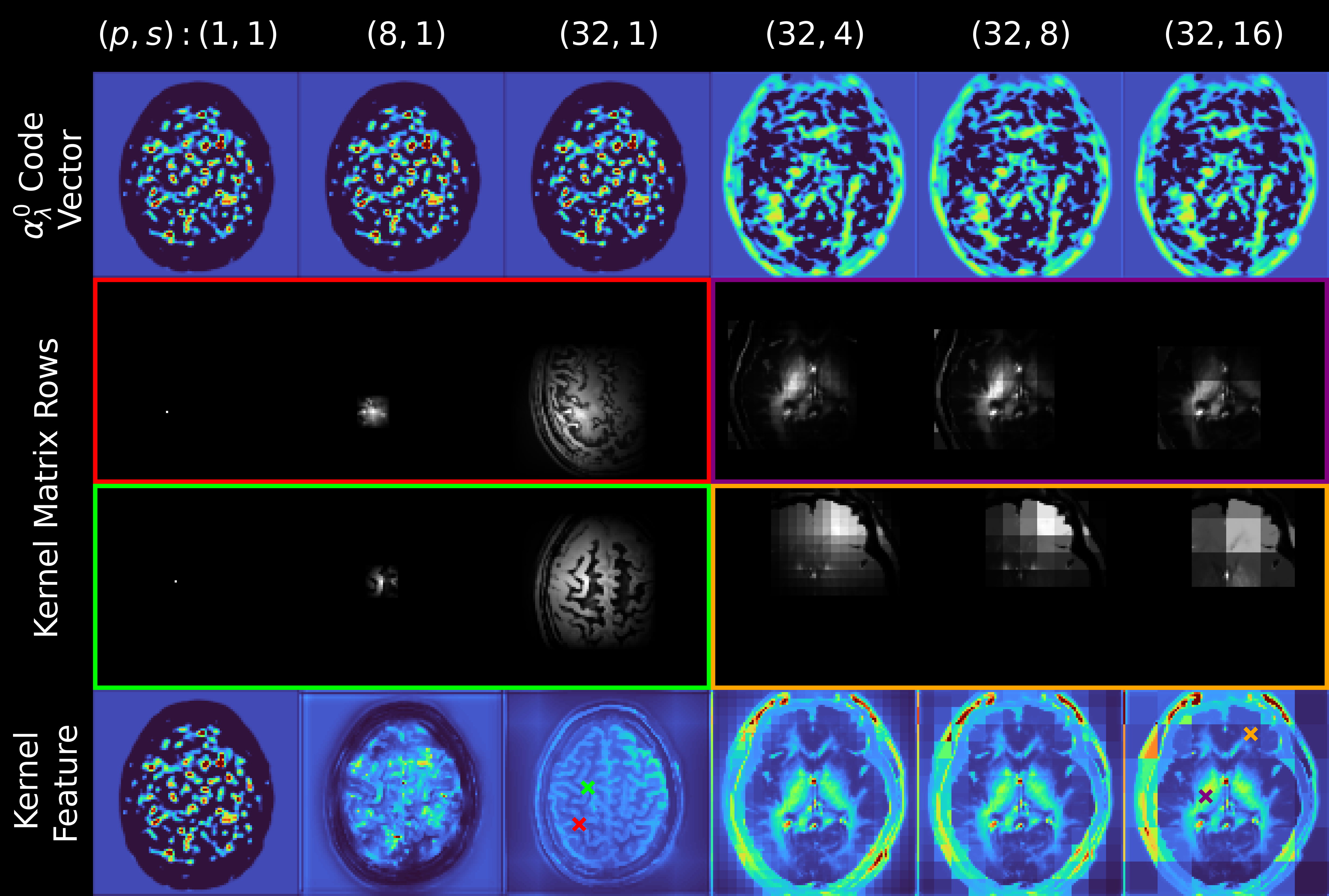}
\caption{Visualisation of latent-space kernel features for an out-of-distribution dose reduction factor of $\times 1000$. Row 1 shows code vectors $\alpha_\lambda^l$ as defined in equation (10) reshaped to image dimensions, rows 2 and 3 show select rows from latent space kernel functions $K_{z,c}^l$ as designated by the green and red cross, and row 4 shows corresponding latent-space kernel features. Increasing the neighbourhood size $p$ as defined in equation (12) provides increased levels of MR guidance. Increasing the stride factor $s$ reduces the modelling capacity of kernel features providing a boost in computational efficiency.}
\end{figure}

\subsubsection{Information constraints}
Figure 13 shows PSNR and SSIM as a function of dose reduction factor for varying values of $\gamma_{max}$. For a given value of $\gamma_{max}$, applying information constraints to increasingly shallow layers increases the strength \color{black} of the regularisation. Constraints applied at a given layer constrain the information content of subsequent layers. Based \color{black} on these empirical results, we nominally apply information maximisation constraints to layer $l=3$ with $\gamma_{max} = 1e^{-2}$ for dose reduction factors $\geq 500$.

Figure 14 shows PSNR and SSIM as a function of dose reduction factor for varying values of $\gamma_{min}$ for paired $^{18}F$-FDG. \color{black} For a given value of $\gamma_{min}$, including increasingly deep kernel features as inputs to the information minimisation constraint increases the strength of the regularisation. Providing more information to the discriminator requires deep PET features to contain less information in order to satisfy the information \color{black} minimisation constraint.

Figure 15 provides a visual comparison highlighting the utility of information constraints in contrast to kernel function patch size. While increasing the kernel function patch size \color{black}provides good anatomical consistency, the information constraint on deep PET layers helps to control the biasing in the putamen\color{black} and caudate regions of the brain for $\times 200$ dose reduction on paired $^{18}F$-FDOPA PET-MR data. This highlights that unconstrained deep features are sensitive to the out-of-distribution noise in the low-dose images, while the kernel features in shallow layers provide an averaging effect which reduces the sensitivity to large noise levels.

\begin{figure}[t!]
\centering
\includegraphics[width=0.49\textwidth]{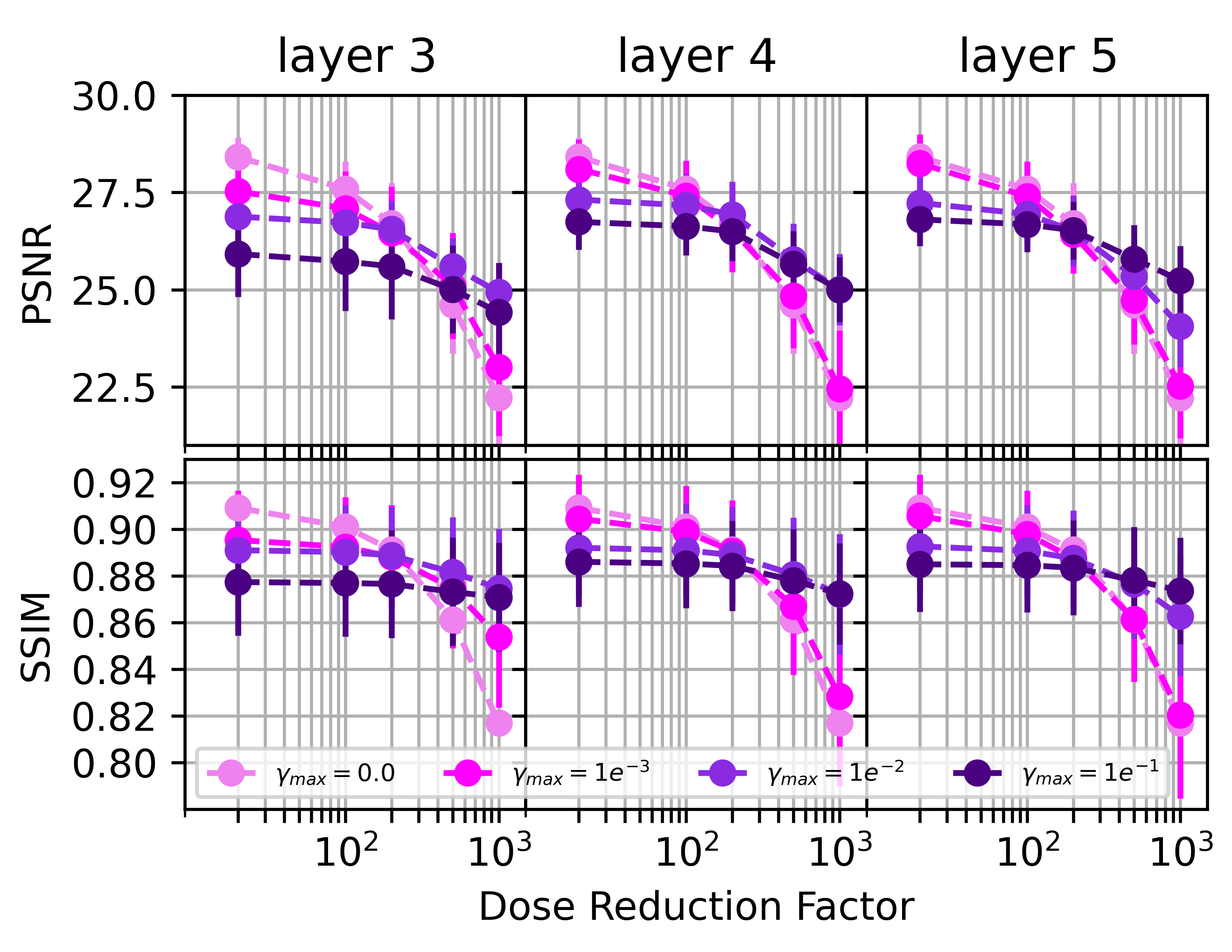}
\caption{Information maximisation constraint applied to increasingly deeper PET encoder features as input to the discriminator network. Kernel functions use patch sizes of $\{32,16,8,4,2,1\}$ for increasingly deep layers and $\gamma_{min} = 0.0$.}
\end{figure}

\begin{figure}[t!]
\centering
\includegraphics[width=0.49\textwidth]{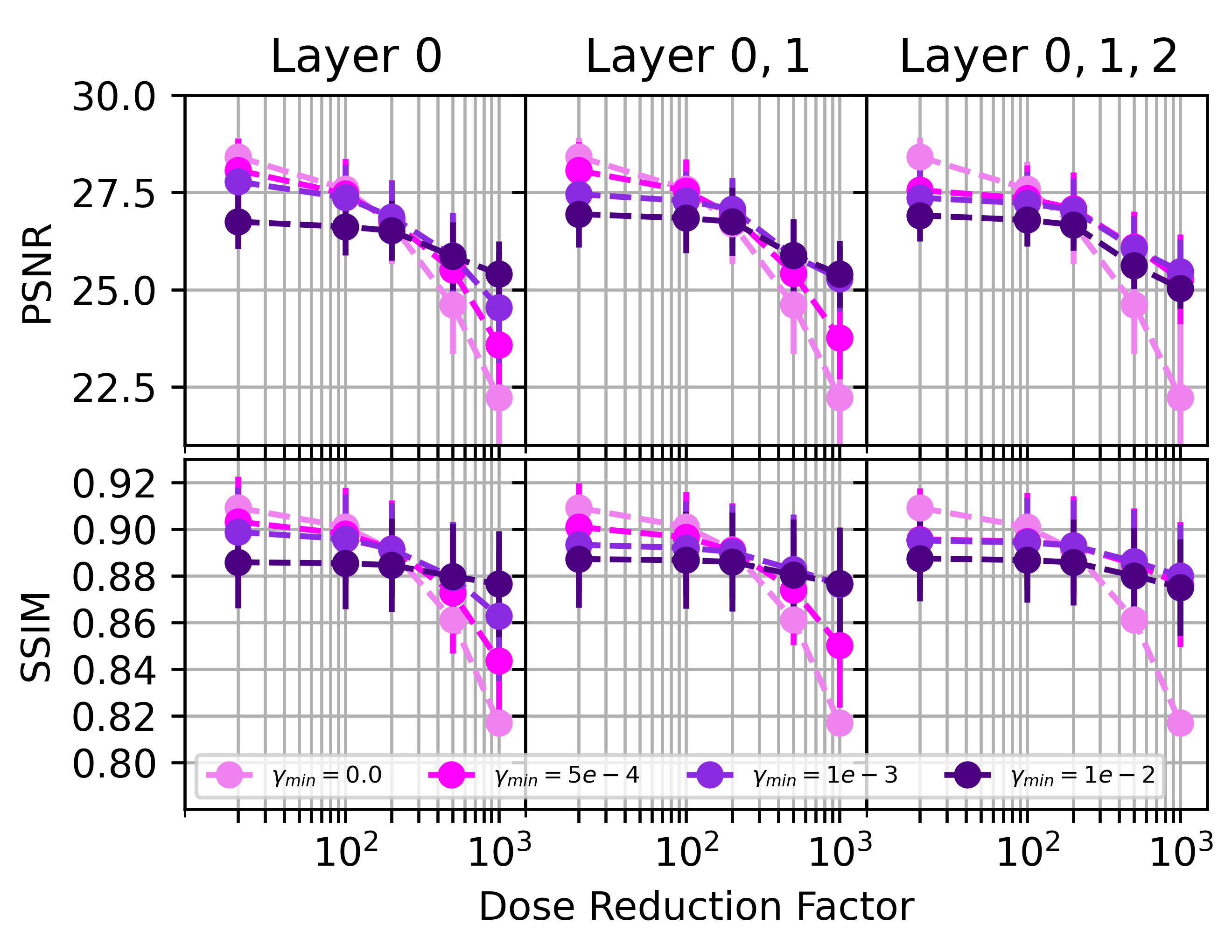}
\caption{Information minimisation constraint including increasingly deeper shallow kernel features as inputs to the discriminator network. Kernel functions use patch sizes of $\{32,16,8,4,2,1\}$ for increasingly deep layers and $\gamma_{max} = 0.0$.}
\end{figure}

\begin{figure}[h!]
\centering
\includegraphics[width=0.49\textwidth]{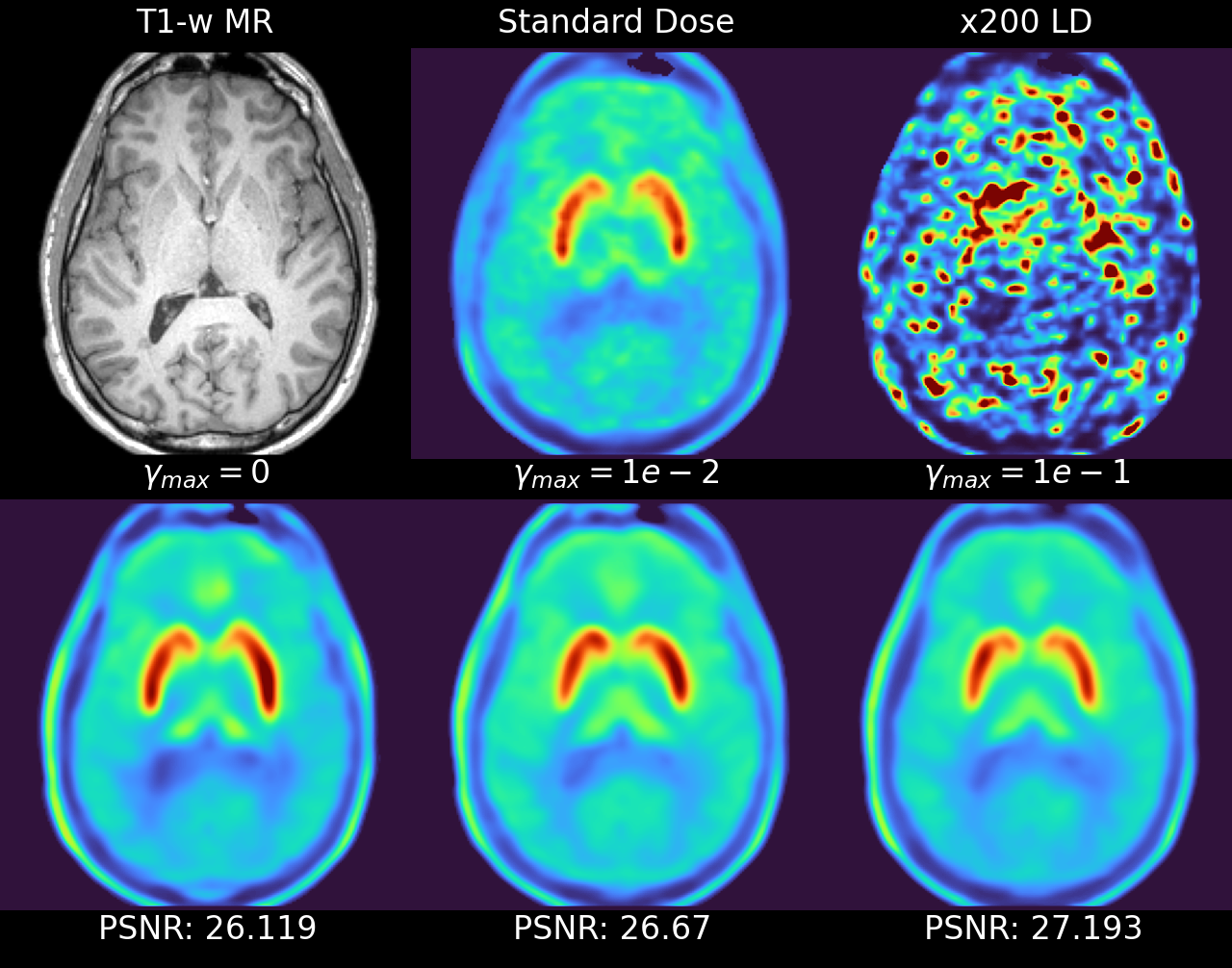}
\caption{Varying information constraint strength for paired $^{18F}$-FDOPA brain at $\times 200$ dose reduction. Large kernel patch size improves anatomical consistency and increasing information constraint strength helps reduce biasing in the caudate and putamen.}
\vspace{-0.4cm}
\end{figure}

\section{Discussion}
The proposed method demonstrates significantly improved performance across out-of-distribution reduction factors for both paired and unpaired PET-MR data. Our method allows for flexibility in the clinic for limiting radiation doses ad hoc, and relaxing margins for error in clinical workflows. The PSNR of measured data varies based on a number of patient-specific factors, including post-injection uptake duration, administered dose and the level of photon attenuation from patient anatomy. Patient specific variations in PSNR are exacerbated in low-dose PET due to the non-linear relationship between acquired counts and the PSNR of the measured signal, likely resulting in cases of drastic under sampling of list mode data in clinical low-dose PET. Conventional supervised learning methods such as the reference methods used in this work, may therefore generalise poorly to a clinical setting where these factors are of importance. Additionally, our method requires no additional data to achieve improved out-of-distribution performance across varying dose levels.

\noindent Although this work focuses on brain imaging, our method may also be applicable to whole body PET imaging. Whole body CT provides rich anatomical contrast throughout the body which may prove useful for applying this method more generally where MR is not readily available and coregistration with template MR may provde more challenging than for the brain. CT also provides contrast between lesions and healthy tissues for a number of cancers, hypothetically providing the ability to apply the proposed method whilst maintaining lesion recovery. Additional experiments need to be performed to test the applicability of this method to varying anatomy and PET tracers. Open source datasets provide readily accessible large cohort data to test the performance on images with pathology, however these generally contain reconstructed standard dose PET images and data from a complimentary imaging modality, rather than the raw data necessary for accurately synthesising low-dose PET images. While low-dose images may be approximated by forward projecting and Poisson sampling standard-dose image data, the realism of low-dose PET images generated by this method is diminished. 

\noindent While the results demonstrated on a cohort of healthy subjects are promising, performance in the presence of pathology must be rigorously evaluated to identify areas where this method can be effectively applied. Pathological features which provide contrast on MR images consistent with contrast on PET images can likely be reproduced in the MR derived kernel functions and PET derived deep kernel coefficients given sufficient training data. Recovery of pathological features which are inconsistent across modalities will likely be challenging at large dose reduction factors. These inconsistencies can arise from a physiological source but also can be caused by MR motion and aliasing artefacts and MR protocol variation. The anatomical information available from PET images can be introduced by reducing the patch size used to learn deep kernel matrices in training, with the limit of a unit patch size resulting in diagonal kernel matrices, producing a latent space representation consistent with a conventional deep learning approach. In this sense, the proposed method does not explicitly address the problem of recovering PET specific pathology, however, it may be integrated into conventional approaches in combination with other methods aiming to solve the problem of recovering PET specific pathology.

\noindent Finally, the implementation presented uses convolutional encoders for each layer, however self-attention mechanisms \cite{Shaw2018self} or vision transformer blocks \cite{Dosovitskiy2020image}\cite{Liang2021_SwinIR}\cite{Liu2021_Swin} could hypothetically be used to potentially further improve in-distribution performance. Similarly, a fully 3D implementation would provide improved performance relative to the 2.5D implementation used in this work. An increasingly large 2.5D or fully 3D implementation would require larger GPU space for storing gradients in memory for the kernel calculations. Training in parallel on multiple GPUs and, if necessary, applying a sufficiently large kernel stride factor can make training and implementation in 3D feasible. Additionally, the forward pass time of our method is unlikely to be a limiting factor in moving towards a 3D implementation, given the training times seen in this work.

\section{Conclusion}
\noindent Kernel representations of latent-space feature maps using paired or unpaired MR provides better generalisability across previously unseen dose reduction factors at inference than conventional deep-learning based approaches, and provides performance comparable to conventional supervised methods when applied to in-distribution dose-reduction factors. Further investigation into the performance on low-dose data with pathological features, the utility of other anatomical imaging modalities and performance of different anatomical regions may demonstrate potential for much broader clinical impact.

\bibliographystyle{ieeetr}
\bibliography{KernelRepresentationsOfDeepLatentFeaturesForLowDosePETMRImagingWithVaryingLevelsOfDoseReduction}

\begin{thebibliography}{10}

\bibitem{Chen2018_PETMR_Review}
Z.~Chen, S.~D. Jamadar, S.~Li, F.~Sforazzini, J.~Baran, N.~Ferris, N.~J. Shah,
  and G.~F. Egan, ``From simultaneous to synergistic mr-pet brain imaging: A
  review of hybrid mr-pet imaging methodologies,'' {\em Human brain mapping},
  vol.~39, no.~12, pp.~5126--5144, 2018.

\bibitem{Wang2014_KernelMethod}
G.~Wang and J.~Qi, ``Pet image reconstruction using kernel method,'' {\em IEEE
  transactions on medical imaging}, vol.~34, no.~1, pp.~61--71, 2014.

\bibitem{8463582}
G.~Wang, ``High temporal-resolution dynamic pet image reconstruction using a
  new spatiotemporal kernel method,'' {\em IEEE Transactions on Medical
  Imaging}, vol.~38, no.~3, pp.~664--674, 2019.

\bibitem{8118149}
K.~Gong, J.~Cheng-Liao, G.~Wang, K.~T. Chen, C.~Catana, and J.~Qi, ``Direct
  patlak reconstruction from dynamic pet data using the kernel method with mri
  information based on structural similarity,'' {\em IEEE Transactions on
  Medical Imaging}, vol.~37, no.~4, pp.~955--965, 2018.

\bibitem{Mehranian2017_Prior}
A.~Mehranian, M.~A. Belzunce, F.~Niccolini, M.~Politis, C.~Prieto,
  F.~Turkheimer, A.~Hammers, and A.~J. Reader, ``Pet image reconstruction using
  multi-parametric anato-functional priors,'' {\em Physics in Medicine and
  Biology}, vol.~62, no.~15, p.~5975, 2017.

\bibitem{Raczynski2020}
L.~Raczy{\'n}ski, W.~Wi{\'s}licki, K.~Klimaszewski, W.~Krzemie{\'n}, P.~Kopka,
  P.~Kowalski, R.~Shopa, M.~Ba{\l}a, J.~Chhokar, C.~Curceanu, {\em et~al.},
  ``3d tof-pet image reconstruction using total variation regularization,''
  {\em Physica Medica}, vol.~80, pp.~230--242, 2020.

\bibitem{Sudarshan2021}
V.~P. Sudarshan, U.~Upadhyay, G.~F. Egan, Z.~Chen, and S.~P. Awate, ``Towards
  lower-dose pet using physics-based uncertainty-aware multimodal learning with
  robustness to out-of-distribution data,'' {\em Medical Image Analysis},
  vol.~73, p.~102187, 2021.

\bibitem{Pain2022_Review}
C.~D. Pain, G.~F. Egan, and Z.~Chen, ``Deep learning-based image reconstruction
  and post-processing methods in positron emission tomography for low-dose
  imaging and resolution enhancement,'' {\em European Journal of Nuclear
  Medicine and Molecular Imaging}, vol.~49, no.~9, pp.~3098--3118, 2022.

\bibitem{Mehranian2022_LD_multicentre}
A.~Mehranian, S.~D. Wollenweber, M.~D. Walker, K.~M. Bradley, P.~A. Fielding,
  K.-H. Su, R.~Johnsen, F.~Kotasidis, F.~P. Jansen, and D.~R. McGowan, ``Image
  enhancement of whole-body oncology [18f]-fdg pet scans using deep neural
  networks to reduce noise,'' {\em European journal of nuclear medicine and
  molecular imaging}, vol.~49, no.~2, pp.~539--549, 2022.

\bibitem{Chang2012_SNR}
T.~Chang, G.~Chang, J.~W. Clark~Jr, R.~H. Diab, E.~Rohren, and O.~R. Mawlawi,
  ``Reliability of predicting image signal-to-noise ratio using noise
  equivalent count rate in pet imaging,'' {\em Medical physics}, vol.~39,
  no.~10, pp.~5891--5900, 2012.

\bibitem{xu2017}
J.~Xu, E.~Gong, J.~Pauly, and G.~Zaharchuk, ``200x low-dose pet reconstruction
  using deep learning,'' {\em arXiv preprint arXiv:1712.04119}, 2017.

\bibitem{Chen2019_M1}
K.~T. Chen, E.~Gong, F.~B. de~Carvalho~Macruz, J.~Xu, A.~Boumis, M.~Khalighi,
  K.~L. Poston, S.~J. Sha, M.~D. Greicius, E.~Mormino, {\em et~al.},
  ``Ultra--low-dose 18f-florbetaben amyloid pet imaging using deep learning
  with multi-contrast mri inputs,'' {\em Radiology}, vol.~290, no.~3,
  pp.~649--656, 2019.

\bibitem{Chen2021_M1}
K.~T. Chen, T.~N. Toueg, M.~E.~I. Koran, G.~Davidzon, M.~Zeineh, D.~Holley,
  H.~Gandhi, K.~Halbert, A.~Boumis, G.~Kennedy, {\em et~al.}, ``True
  ultra-low-dose amyloid pet/mri enhanced with deep learning for clinical
  interpretation,'' {\em European journal of nuclear medicine and molecular
  imaging}, vol.~48, pp.~2416--2425, 2021.

\bibitem{Ouyang2019_M1}
J.~Ouyang, K.~T. Chen, E.~Gong, J.~Pauly, and G.~Zaharchuk, ``Ultra-low-dose
  pet reconstruction using generative adversarial network with feature matching
  and task-specific perceptual loss,'' {\em Medical physics}, vol.~46, no.~8,
  pp.~3555--3564, 2019.

\bibitem{Zhao2020_M2}
K.~Zhao, L.~Zhou, S.~Gao, X.~Wang, Y.~Wang, X.~Zhao, H.~Wang, K.~Liu, Y.~Zhu,
  and H.~Ye, ``Study of low-dose pet image recovery using supervised learning
  with cyclegan,'' {\em Plos one}, vol.~15, no.~9, p.~e0238455, 2020.

\bibitem{Zhou2020_M2}
L.~Zhou, J.~D. Schaefferkoetter, I.~W. Tham, G.~Huang, and J.~Yan, ``Supervised
  learning with cyclegan for low-dose fdg pet image denoising,'' {\em Medical
  Image Analysis}, vol.~65, p.~101770, 2020.

\bibitem{Gong2019_emnet}
K.~Gong, D.~Wu, K.~Kim, J.~Yang, G.~El~Fakhri, Y.~Seo, and Q.~Li, ``Emnet: an
  unrolled deep neural network for pet image reconstruction,'' in {\em Medical
  imaging 2019: Physics of medical imaging}, vol.~10948, pp.~1203--1208, SPIE,
  2019.

\bibitem{Mehranian2019_FBSEM}
A.~Mehranian and A.~J. Reader, ``Model-based deep learning pet image
  reconstruction using forward-backward splitting expectation maximisation,''
  in {\em 2019 IEEE Nuclear Science Symposium and Medical Imaging Conference
  (NSS/MIC)}, pp.~1--4, 2019.

\bibitem{Gong2018_iterative}
K.~Gong, J.~Guan, K.~Kim, X.~Zhang, J.~Yang, Y.~Seo, G.~El~Fakhri, J.~Qi, and
  Q.~Li, ``Iterative pet image reconstruction using convolutional neural
  network representation,'' {\em IEEE transactions on medical imaging},
  vol.~38, no.~3, pp.~675--685, 2018.

\bibitem{Ulyanov2018_DIP}
D.~Ulyanov, A.~Vedaldi, and V.~Lempitsky, ``Deep image prior,'' in {\em
  Proceedings of the IEEE conference on computer vision and pattern
  recognition}, pp.~9446--9454, 2018.

\bibitem{Cui2019_DIPImage}
J.~Cui, K.~Gong, N.~Guo, C.~Wu, X.~Meng, K.~Kim, K.~Zheng, Z.~Wu, L.~Fu, B.~Xu,
  {\em et~al.}, ``Pet image denoising using unsupervised deep learning,'' {\em
  European journal of nuclear medicine and molecular imaging}, vol.~46,
  pp.~2780--2789, 2019.

\bibitem{Sun2021_DIPImage}
H.~Sun, L.~Peng, H.~Zhang, Y.~He, S.~Cao, and L.~Lu, ``Dynamic pet image
  denoising using deep image prior combined with regularization by denoising,''
  {\em IEEE Access}, vol.~9, pp.~52378--52392, 2021.

\bibitem{Hashimoto_2021_4DDIP}
F.~Hashimoto, H.~Ohba, K.~Ote, A.~Kakimoto, H.~Tsukada, and Y.~Ouchi, ``4d deep
  image prior: dynamic pet image denoising using an unsupervised
  four-dimensional branch convolutional neural network,'' {\em Physics in
  Medicine and Biology}, vol.~66, p.~015006, jan 2021.

\bibitem{Onishi_DIP_Pretrain}
Y.~Onishi, F.~Hashimoto, K.~Ote, K.~Matsubara, and M.~Ibaraki,
  ``Self-supervised pre-training for deep image prior-based robust pet image
  denoising,'' {\em IEEE Transactions on Radiation and Plasma Medical
  Sciences}, pp.~1--1, 2023.

\bibitem{Cui2021_population}
J.~Cui, K.~Gong, N.~Guo, C.~Wu, K.~Kim, H.~Liu, and Q.~Li, ``Populational and
  individual information based pet image denoising using conditional
  unsupervised learning,'' {\em Physics in Medicine and Biology}, vol.~66,
  no.~15, p.~155001, 2021.

\bibitem{Gong2018_DIPRecon}
K.~Gong, C.~Catana, J.~Qi, and Q.~Li, ``Pet image reconstruction using deep
  image prior,'' {\em IEEE transactions on medical imaging}, vol.~38, no.~7,
  pp.~1655--1665, 2018.

\bibitem{Yokota_2019_ICCV}
T.~Yokota, K.~Kawai, M.~Sakata, Y.~Kimura, and H.~Hontani, ``Dynamic pet image
  reconstruction using nonnegative matrix factorization incorporated with deep
  image prior,'' in {\em Proceedings of the IEEE/CVF International Conference
  on Computer Vision (ICCV)}, October 2019.

\bibitem{Hashimoto_2022_DIPRecon}
F.~Hashimoto, K.~Ote, and Y.~Onishi, ``Pet image reconstruction incorporating
  deep image prior and a forward projection model,'' {\em IEEE Transactions on
  Radiation and Plasma Medical Sciences}, vol.~6, no.~8, pp.~841--846, 2022.

\bibitem{Shan2023_PartialDataDIPRecon}
Q.~Shan, J.~Wang, and D.~Liu, ``Deep image prior based pet reconstruction from
  partial data,'' {\em IEEE Transactions on Radiation and Plasma Medical
  Sciences}, pp.~1--1, 2023.

\bibitem{Ote2023_DIPlistmode}
K.~Ote, F.~Hashimoto, Y.~Onishi, T.~Isobe, and Y.~Ouchi, ``List-mode pet image
  reconstruction using deep image prior,'' {\em IEEE Transactions on Medical
  Imaging}, 2023.

\bibitem{Wavelet_Kernel}
Z.~Ashouri, G.~Wang, R.~M. Dansereau, and R.~A. deKemp, ``Evaluation of wavelet
  kernel-based pet image reconstruction,'' {\em IEEE Transactions on Radiation
  and Plasma Medical Sciences}, vol.~6, no.~5, pp.~564--573, 2022.

\bibitem{10026307}
S.~Guo, Y.~Sheng, L.~Chai, and J.~Zhang, ``Pet image reconstruction with kernel
  and kernel space composite regularizer,'' {\em IEEE Transactions on Medical
  Imaging}, vol.~42, no.~6, pp.~1786--1798, 2023.

\bibitem{NeuralKEM}
S.~Li, K.~Gong, R.~D. Badawi, E.~J. Kim, J.~Qi, and G.~Wang, ``Neural kem: A
  kernel method with deep coefficient prior for pet image reconstruction,''
  {\em IEEE Transactions on Medical Imaging}, vol.~42, no.~3, pp.~785--796,
  2023.

\bibitem{DeepKernelRep}
S.~Li and G.~Wang, ``Deep kernel representation for image reconstruction in
  pet,'' {\em IEEE Transactions on Medical Imaging}, vol.~41, no.~11,
  pp.~3029--3038, 2022.

\bibitem{Hofmann2008_KernelMethodsInMachineLearning}
T.~Hofmann, B.~Sch{\"o}lkopf, and A.~J. Smola, ``Kernel methods in machine
  learning,'' 2008.

\bibitem{Hashimoto2021_DIP}
F.~Hashimoto, H.~Ohba, K.~Ote, A.~Kakimoto, H.~Tsukada, and Y.~Ouchi, ``4d deep
  image prior: dynamic pet image denoising using an unsupervised
  four-dimensional branch convolutional neural network,'' {\em Physics in
  Medicine and Biology}, vol.~66, no.~1, p.~015006, 2021.

\bibitem{Gong2018_DIP}
K.~Gong, C.~Catana, J.~Qi, and Q.~Li, ``Pet image reconstruction using deep
  image prior,'' {\em IEEE transactions on medical imaging}, vol.~38, no.~7,
  pp.~1655--1665, 2018.

\bibitem{hjelm2019_DeepInfoMax}
D.~Hjelm, A.~Fedorov, S.~Lavoie-Marchildon, K.~Grewal, P.~Bachman,
  A.~Trischler, and Y.~Bengio, ``Learning deep representations by mutual
  information estimation and maximization,'' in {\em ICLR 2019}, ICLR, April
  2019.

\bibitem{ANTS2009}
B.~B. Avants, N.~Tustison, G.~Song, {\em et~al.}, ``Advanced normalization
  tools (ants),'' {\em Insight j}, vol.~2, no.~365, pp.~1--35, 2009.

\bibitem{Jenkinson2012_FSL}
M.~Jenkinson, C.~F. Beckmann, T.~E. Behrens, M.~W. Woolrich, and S.~M. Smith,
  ``Fsl,'' {\em Neuroimage}, vol.~62, no.~2, pp.~782--790, 2012.

\bibitem{Sudarshan2021_Bowsher}
V.~P. Sudarshan, S.~Li, S.~D. Jamadar, G.~F. Egan, S.~P. Awate, and Z.~Chen,
  ``Incorporation of anatomical mri knowledge for enhanced mapping of brain
  metabolism using functional pet,'' {\em NeuroImage}, vol.~233, p.~117928,
  2021.

\bibitem{STIR}
K.~Thielemans, C.~Tsoumpas, S.~Mustafovic, T.~Beisel, P.~Aguiar, N.~Dikaios,
  and M.~W. Jacobson, ``Stir: software for tomographic image reconstruction
  release 2,'' {\em Physics in Medicine and Biology}, vol.~57, no.~4, p.~867,
  2012.

\bibitem{Shaw2018self}
P.~Shaw, J.~Uszkoreit, and A.~Vaswani, ``Self-attention with relative position
  representations,'' {\em arXiv preprint arXiv:1803.02155}, 2018.

\bibitem{Dosovitskiy2020image}
A.~Dosovitskiy, L.~Beyer, A.~Kolesnikov, D.~Weissenborn, X.~Zhai,
  T.~Unterthiner, M.~Dehghani, M.~Minderer, G.~Heigold, S.~Gelly, {\em et~al.},
  ``An image is worth 16x16 words: Transformers for image recognition at
  scale,'' {\em arXiv preprint arXiv:2010.11929}, 2020.

\bibitem{Liang2021_SwinIR}
J.~Liang, J.~Cao, G.~Sun, K.~Zhang, L.~V. Gool, and R.~Timofte, ``Swinir: Image
  restoration using swin transformer,'' 2021.

\bibitem{Liu2021_Swin}
Z.~Liu, Y.~Lin, Y.~Cao, H.~Hu, Y.~Wei, Z.~Zhang, S.~Lin, and B.~Guo, ``Swin
  transformer: Hierarchical vision transformer using shifted windows,'' in {\em
  Proceedings of the IEEE/CVF international conference on computer vision},
  pp.~10012--10022, 2021.

\end{thebibliography}

\end{document}